\documentclass{article}
\usepackage{authblk}
\usepackage[utf8]{inputenc}
\usepackage{main}
\usepackage{microtype}
\usepackage{subcaption}
\usepackage{graphicx}
\usepackage{times}
\usepackage{latexsym}
\usepackage{amsmath}
\usepackage{float}
\usepackage{footnote}
\usepackage{enumitem}
\usepackage{bm}
\usepackage{arydshln}
\usepackage{booktabs}
\usepackage{array}     
\usepackage{multicol}
\usepackage{multirow}
\usepackage{color}
\usepackage{xcolor}     
\usepackage{colortbl}
\usepackage{bbding}
\usepackage{makecell}
\usepackage{mathtools}
\usepackage{imakeidx}
\usepackage{longtable}
\usepackage{wrapfig}
\usepackage{rotating}
\makeindex
\usepackage{arydshln}
\usepackage{lipsum}
\usepackage{natbib}
\usepackage[toc]{multitoc}
\usepackage[edges]{forest}
\usepackage[normalem]{ulem}
\definecolor{mydarkblue}{rgb}{0,0.08,0.45}
\usepackage[colorlinks=true,linkcolor=mydarkblue,citecolor=mydarkblue,filecolor=mydarkblue,urlcolor=mydarkblue]{hyperref}
\usepackage{CJKutf8}
\usepackage{awesomebox} 
\usepackage{bbding}
\usepackage[most]{tcolorbox}
\usepackage{booktabs}
\usepackage{geometry}
\definecolor{wkblue}{rgb}{0.2, 0.3, 0.6}
\definecolor{meta-color}{rgb}{0.5, 0.5, 0.5}
\usepackage{amsmath}
\usepackage{enumitem}
\usepackage{lscape} 
\usepackage{booktabs}
\usepackage{algorithm}    

\usepackage{algorithmicx}
\usepackage{algpseudocode}
\usepackage{tabularx,booktabs}
\usepackage{makecell}
\usepackage{pifont}
\usepackage{tcolorbox}
\usepackage{amssymb}
\usepackage{amsfonts}

\usepackage{multirow}

\usepackage[tikz]{bclogo}
\usepackage[framemethod=tikz]{mdframed}
\definecolor{bgblue}{RGB}{245,243,253}
\definecolor{ttblue}{RGB}{91,194,224}

\mdfdefinestyle{mystyle}{%
  rightline=true,
  innerleftmargin=10,
  innerrightmargin=10,
  outerlinewidth=3pt,
  topline=false,
  rightline=true,
  bottomline=false,
  skipabove=\topsep,
  skipbelow=\topsep
}

\newtcolorbox{myboxi}[1][]{
  breakable,
  title=#1,
  colback=red!5,
  colbacktitle=red!5,
  coltitle=black,
  fonttitle=\bfseries,
  bottomrule=0pt,
  toprule=0pt,
  leftrule=2pt,
  rightrule=2pt,
  titlerule=0pt,
  arc=0pt,
  outer arc=0pt,
  colframe=red,
}

\newtcolorbox{myboxnote}[1][]{
  breakable,
  title=#1,
  colback=orange!0,
  colbacktitle=orange!0,
  coltitle=black,
  fonttitle=\bfseries,
  bottomrule=0pt,
  toprule=0pt,
  leftrule=2pt,
  rightrule=2pt,
  titlerule=0pt,
  arc=0pt,
  outer arc=0pt,
  colframe=orange,
}

\newtcolorbox{myboxii}[1][]{
  breakable,
  freelance,
  title=#1,
  colback=white,
  colbacktitle=white,
  coltitle=black,
  fonttitle=\bfseries,
  bottomrule=0pt,
  boxrule=0pt,
  colframe=white,
  overlay unbroken and first={
  \draw[red!75!black,line width=3pt]
    ([xshift=5pt]frame.north west) -- 
    (frame.north west) -- 
    (frame.south west);
  \draw[red!75!black,line width=3pt]
    ([xshift=-5pt]frame.north east) -- 
    (frame.north east) -- 
    (frame.south east);
  },
  overlay unbroken app={
  \draw[red!75!black,line width=3pt,line cap=rect]
    (frame.south west) -- 
    ([xshift=5pt]frame.south west);
  \draw[red!75!black,line width=3pt,line cap=rect]
    (frame.south east) -- 
    ([xshift=-5pt]frame.south east);
  },
  overlay middle and last={
  \draw[red!75!black,line width=3pt]
    (frame.north west) -- 
    (frame.south west);
  \draw[red!75!black,line width=3pt]
    (frame.north east) -- 
    (frame.south east);
  },
  overlay last app={
  \draw[red!75!black,line width=3pt,line cap=rect]
    (frame.south west) --
    ([xshift=5pt]frame.south west);
  \draw[red!75!black,line width=3pt,line cap=rect]
    (frame.south east) --
    ([xshift=-5pt]frame.south east);
  },
}

\usepackage{fancyhdr} 
\usepackage{blindtext} 
\usepackage{makecell}

\DeclareCaptionFont{black}{\color{black}}

\definecolor{myblue}{rgb}{0.9, 0.1, 0.94}
\definecolor{mygreen}{rgb}{0.64, 0.56, 0.88}
\definecolor{myyellow}{rgb}{0.68, 0.6, 0.1}
\definecolor{fancygreen}{rgb}{0.33, 0.68, 0.20}
\definecolor{salmon}{rgb}{0.94, 0.52, 0.49}
\definecolor{tablegreen}{rgb}{0.82, 0.94, 0.75}
\definecolor{tableblue}{rgb}{0.81, 0.90, 0.94}
\definecolor{tablered}{rgb}{0.97, 0.85, 0.85}
\definecolor{tableorange}{rgb}{0.96, 0.85, 0.81}

\newenvironment{itemize*}%
 {\leftmargini=10pt\begin{itemize}%
  \setlength{\itemsep}{0pt}%
  \setlength{\parskip}{0pt}%
  }%
 {\end{itemize}}
\newenvironment{enumerate*}%
 {\begin{enumerate}%
  \setlength{\itemsep}{0pt}%
  \setlength{\parskip}{0pt}}%
 {\end{enumerate}}

\usepackage{xcolor}
\usepackage{listings}

\newcommand\JSONnumbervaluestyle{\color{blue}}
\newcommand\JSONstringvaluestyle{\color{red}}

\newif\ifcolonfoundonthisline

\makeatletter

\lstdefinestyle{json}
{
  showstringspaces    = false,
  keywords            = {false,true},
  alsoletter          = 0123456789.,
  morestring          = [s]{"}{"},
  stringstyle         = \ifcolonfoundonthisline\JSONstringvaluestyle\fi,
  MoreSelectCharTable =%
    \lst@DefSaveDef{`:}\colon@json{\processColon@json},
  basicstyle          = \ttfamily,
  keywordstyle        = \ttfamily\bfseries,
}

\newcommand\processColon@json{%
  \colon@json%
  \ifnum\lst@mode=\lst@Pmode%
    \global\colonfoundonthislinetrue%
  \fi
}

\lst@AddToHook{Output}{%
  \ifcolonfoundonthisline%
    \ifnum\lst@mode=\lst@Pmode%
      \def\lst@thestyle{\JSONnumbervaluestyle}%
    \fi
  \fi
  \lsthk@DetectKeywords%
}

\lst@AddToHook{EOL}%
  {\global\colonfoundonthislinefalse}

\makeatother

\usepackage{etoolbox}
\usepackage{natbib}
\usepackage{url}
\newcounter{bibcount}
\makeatletter
\patchcmd{\@lbibitem}{\item[}{\item[\hfil\stepcounter{bibcount}{[\thebibcount]}}{}{}
\renewcommand\NAT@bibsetup%
  [1]{\setlength{\leftmargin}{\bibhang}\setlength{\itemindent}{-\parindent}%
      \setlength{\itemsep}{\bibsep}\setlength{\parsep}{\z@}}
\makeatother

\definecolor{mybrown}{RGB}{128,64,0}

\definecolor{titlecolor}{HTML}{4c9cff}

\usepackage{diagbox}
\usepackage{xcolor}

\tcbset{colback=cyan!5!white, colframe=cyan!75!black, 
        boxrule=0.5mm, arc=4mm, auto outer arc, 
        width=0.95\textwidth, 
        before=,after=\par\bigskip}

\newcommand{\scrl}{DeepResearcher\xspace}

\begin{document}

\title{DeepResearcher: Scaling Deep Research via Reinforcement Learning in Real-world Environments}

\author{
\textbf{Yuxiang Zheng}$^{1,2,3}$\thanks{Co-first authors} \quad
\textbf{Dayuan Fu}$^{2,3*}$ \quad
\textbf{Xiangkun Hu}$^{2*}$ \\
\textbf{Xiaojie Cai}$^{1,3}$ \quad
\textbf{Lyumanshan Ye}$^{1,3}$ \quad
\textbf{Pengrui Lu}$^{1,3}$ \quad
\textbf{Pengfei Liu}$^{1,2,3}$\thanks{Corresponding author}\\\;\\
\textsuperscript{1}SJTU\quad
\textsuperscript{2}SII\quad
\textsuperscript{3}GAIR
}

\maketitle
\thispagestyle{fancy}

\fancyhead{}
\lhead{\includegraphics[height=0.67cm]{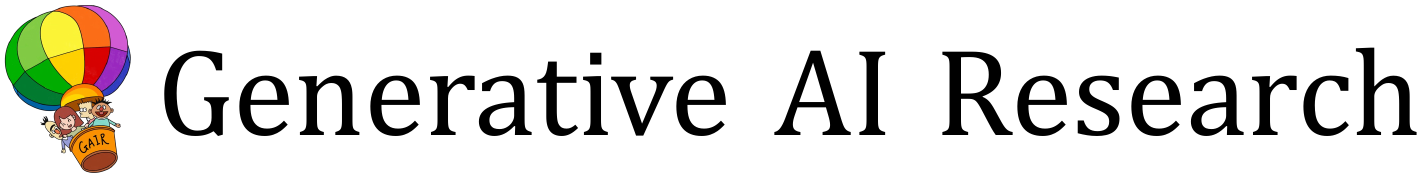}}
\renewcommand{\headrulewidth}{0pt}
\setlength{\headsep}{0mm}

\begin{abstract}

Large Language Models (LLMs) equipped with web search capabilities have demonstrated impressive potential for deep research tasks. However, current approaches predominantly rely on either manually engineered prompts (\emph{prompt engineering-based}) with brittle performance or reinforcement learning within controlled Retrieval-Augmented Generation (RAG) environments (\emph{RAG-based}) that fail to capture the complexities of real-world interaction.
In this paper, we introduce \textbf{\scrl}, the first comprehensive framework for \textbf{end-to-end training} of LLM-based deep research agents through \textbf{scaling reinforcement learning (RL) in real-world environments} with authentic web search interactions. Unlike RAG-based approaches that assume all necessary information exists within a fixed corpus, our method trains agents to navigate the noisy, unstructured, and dynamic nature of the open web. We implement a specialized multi-agent architecture where browsing agents extract relevant information from various webpage structures and overcoming significant technical challenges.
Extensive experiments on open-domain research tasks demonstrate that \scrl achieves substantial improvements of up to \textbf{28.9} points over prompt engineering-based baselines and up to \textbf{7.2} points over RAG-based RL agents. Our qualitative analysis reveals emergent \textbf{cognitive behaviors} from end-to-end RL training, including the ability to formulate plans, cross-validate information from multiple sources, engage in self-reflection to redirect research, and maintain honesty when unable to find definitive answers.
Our results highlight that end-to-end training in real-world web environments is not merely an implementation detail but a fundamental requirement for developing robust research capabilities aligned with real-world applications. We release \scrl at \url{https://github.com/GAIR-NLP/DeepResearcher}.

\end{abstract}

\begin{figure}[h]
    \centering
    \includegraphics[width=0.8\linewidth]{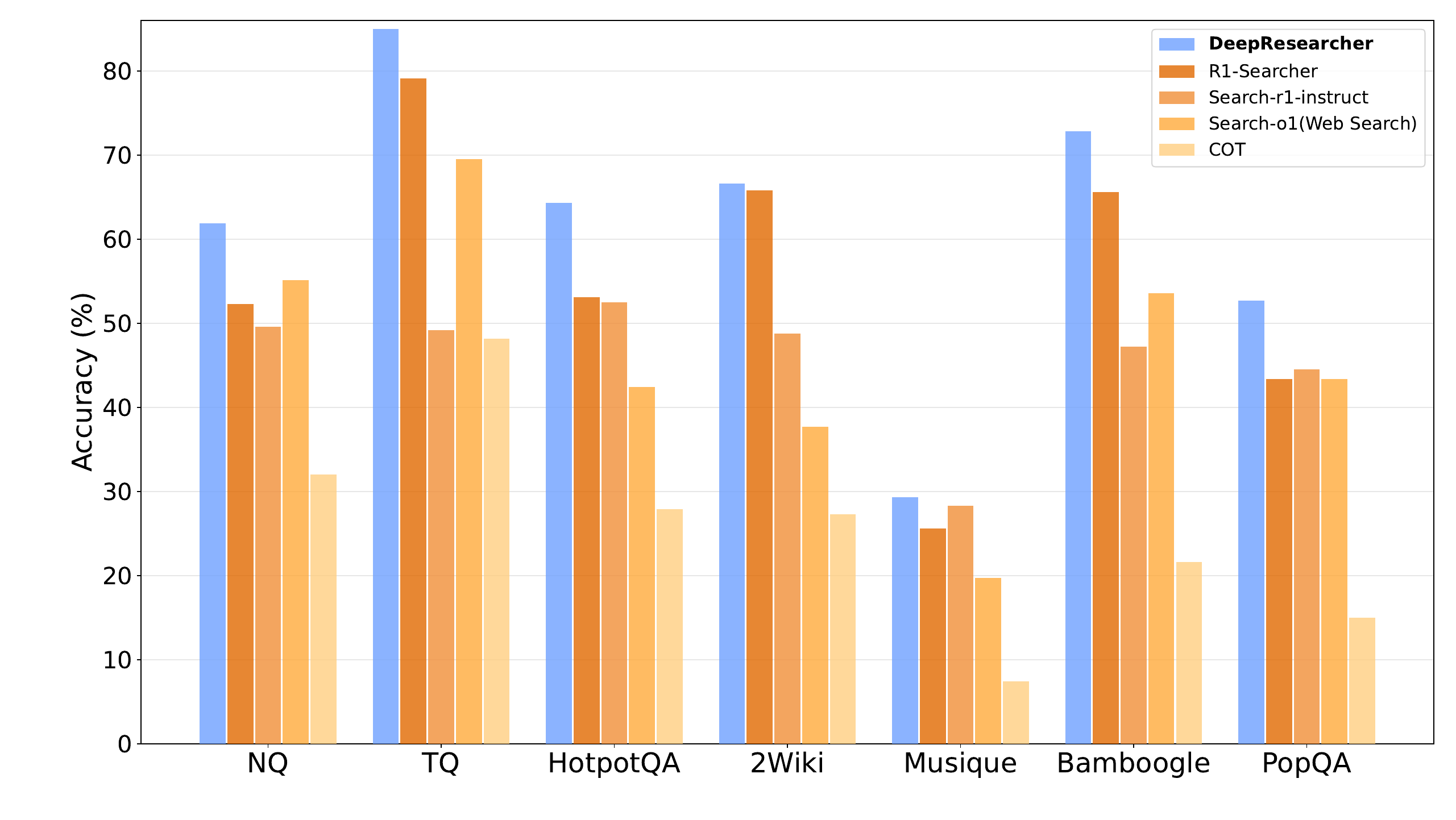}
    \caption{DeepResearcher performs the best on all 7 datasets measured by reliable model-based evaluation.}
    \label{fig:performance}
\end{figure}

\newpage

\pagestyle{fancy}
\lhead{\rightmark}
\renewcommand{\headrulewidth}{0.7pt}
\setlength{\headsep}{5mm}

\begin{figure}[h]
    \centering
    \includegraphics[width=\linewidth]{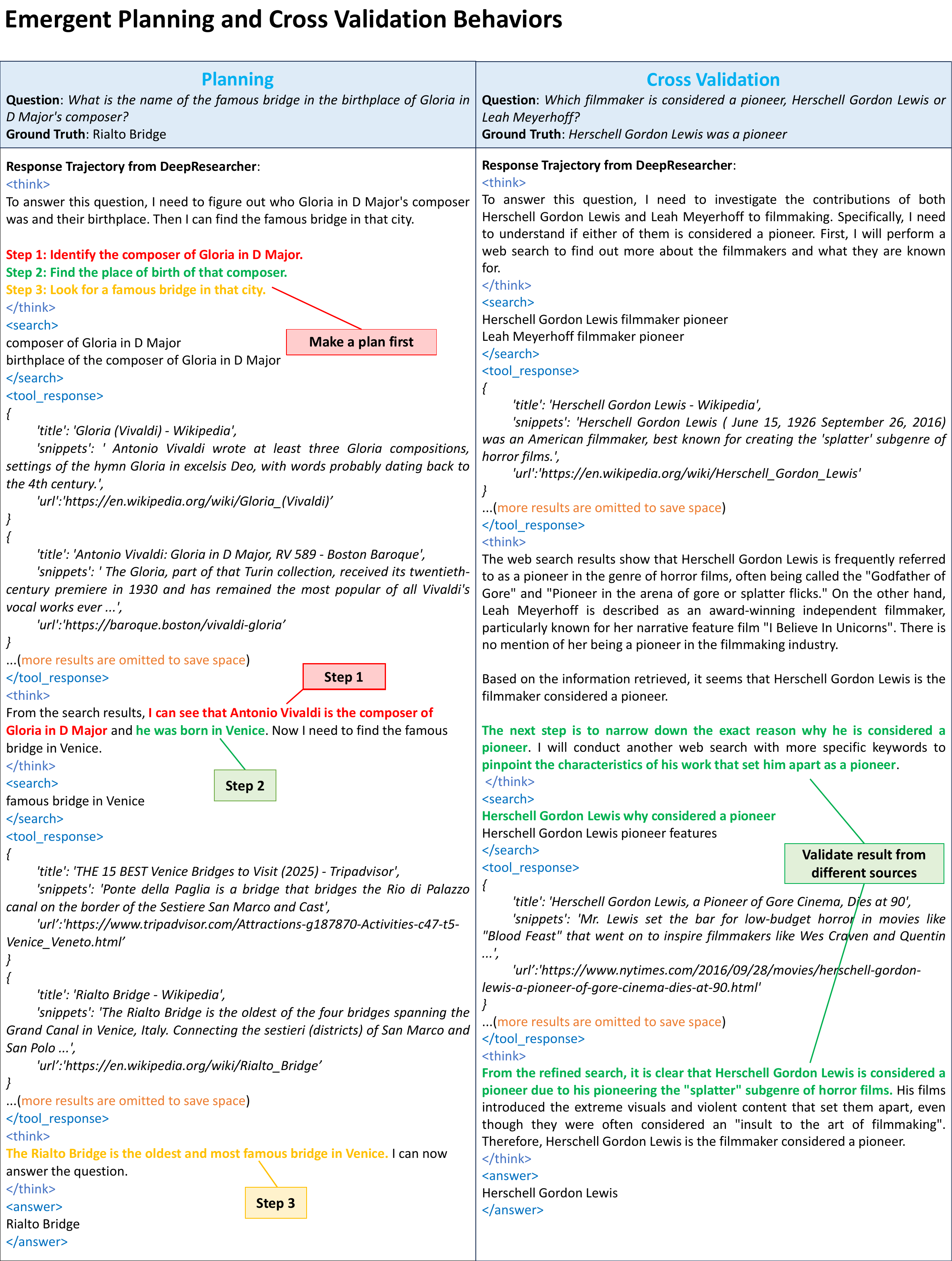}
    \caption{Planning and Cross Validation Behaviors. \textbf{Left}: \scrl is capable of making plans and dynamically adjusting it throughout the reasoning process. Notably, the model can merge steps when appropriate. \textbf{Right}: \scrl identifies the correct answer during its first tool call. However, rather than immediately committing to this result, it proceeds to verify its accuracy through subsequent steps. This rigorous behavior enhances the reliability of the model's responses,  ensuring greater robustness in its final answers. Note: In the actual model, the tool call and tool response format is a JSON string, rather than a tag. However, for clarity and ease of presentation, we have used tags in the figure to represent the output.}
    \label{fig:planning-label}
\end{figure}

\clearpage

\newpage

\section{Introduction}

The emergence of Large Language Models (LLMs) has fundamentally transformed the landscape of artificial intelligence, enabling increasingly autonomous problem-solving capabilities. When equipped with external tools such as web search and code execution~\cite{li2025torlscalingtoolintegratedrl}, these models can tackle complex research tasks that previously required significant human workload and expertise. Notable examples include Gemini and OpenAI Deep Research~\cite{google2024gemini, openai2025deep}, Grok3's DeeperSearch~\cite{xai2025grok}, and open-source projects like MetaGPT~\cite{hong2024metagpt}, OpenManus~\cite{openmanus2025}, and OWL agents~\cite{owl2025}. These systems demonstrate promising capabilities in synthesizing information, writing and executing code, and conducting iterative investigations across diverse domains.
Despite their potential, most current agents are prompt-engineered LLM agents that face significant limitations, while the technical details of commercial systems like OpenAI Deep Research remain completely opaque. Specifically, prompt-engineered agents follow pre-defined workflows designed by developers~\cite{anthropic2024building}, resulting in strict behavioral patterns. Consequently, they often struggle with instruction following, consistent reasoning, exhibit poor generalization to novel tasks, and require extensive manual prompt engineering to achieve reliable performance~\cite{pan2025why}. The brittle nature of these systems becomes particularly evident when confronted with complex, multi-step research scenarios requiring adaptive information gathering through web search. While impressive commercial products exist, reproducible frameworks for creating robust research agents remain elusive.

Recent advances suggest that reinforcement learning (RL) offers a promising path forward for improving LLM capabilities. Studies by \citet{guo2025deepseek} and \citet{team2025kimi} demonstrate that scaling reinforcement learning for LLMs on math and coding tasks~\cite{li2025limr} substantially improves their reasoning abilities. For open-domain tasks, OpenAI has acknowledged using reinforcement learning techniques to enhance their Deep Research agent's capabilities, but detailed methodologies remain proprietary and undisclosed, creating a significant gap in open research. Current open-source efforts to integrate RL with information retrieval, such as Search-R1~\cite{jin2025search}, R1-Searcher~\cite{song2025r1}, and ReSearch~\cite{chen2025research}, have primarily focused on Retrieval-Augmented Generation (RAG) using \emph{static}, \emph{local} text corpora. While these approaches provide valuable insights, they \textbf{fundamentally fail to capture the dynamic, unpredictable nature of real-world web search environments}. The controlled RAG setting operates in a highly sanitized environment with a critical limiting assumption: \emph{that all necessary information already exists within their fixed knowledge base}. \emph{This assumption breaks down in real-world scenarios} where information might be absent, outdated, or require synthesis across domains not covered in the initial knowledge base. Beyond this fundamental limitation, RAG systems also fail to account for the substantial noise, variability in search quality, and the challenges of navigating diverse web content formats and structures.

In this work, we present the first comprehensive study of RL scaling for LLM agents operating with real-world web search capabilities. Our approach, \textbf{\scrl}, trains agents to \textbf{interact directly with live search engines}, thereby learning to handle the inherent variability and complexity of the open web. By \textbf{training in genuine web environments rather than controlled simulations}, our system develops robust capabilities for handling the unpredictable nature of real-world information retrieval and synthesis.

\scrl diverges significantly from prompt-based and RAG-based methods by applying several critical techniques absent from previous work:

\begin{itemize*}
\item \textbf{Scaling RL for Deep Research}: Unlike prompt and SFT-based methods, we directly scale RL training for deep research with only outcome rewards.
\item \textbf{Real-world Environment}: Unlike controlled RAG environments, real web search presents noisy, unstructured, and heterogeneous information sources that require sophisticated filtering and relevance assessment capabilities.
\item \textbf{End-to-end Training}: We train the model end-to-end without human priors, enabling the agent to discover its own problem-solving strategies. This end-to-end approach significantly departs from human-designed workflows.
\item \textbf{Addressing Implementation Challenges}: Training with real web search introduces unique challenges absent in RAG settings, including managing search API rate limits, handling network latency, addressing anti-crawling mechanisms, and processing diverse webpage structures.
\item \textbf{Multi-agent Framework}: Our approach employs a specialized multi-agent architecture where dedicated browsing agents extract relevant information from entire webpages—a stark contrast to RAG-based systems that simply retrieve and present pre-processed text passages.
\end{itemize*}

Our results show that \scrl achieves up to 28.9 points of improvement in research task completion compared to prompt-engineered agents. When compared to RAG-based RL agents, \scrl demonstrates an improvement of up to 7.2 points. These findings suggest that direct interaction with real search environments is not merely an implementation detail but a crucial component for developing robust research capabilities in autonomous systems that can perform effectively in real-world applications.

Furthermore, our qualitative analysis revealed several important cognitive behaviors that emerge from \scrl's end-to-end RL scaling. During problem-solving, \scrl demonstrates abilities to \textbf{make plans} initially, \textbf{cross-validate answers} from multiple sources, engage in \textbf{reflection} to redirect research, and \textbf{maintain honesty} when unable to find exact answers. These capabilities represent important characteristics for deep research agents and mirror skills valued in human researchers.

To conclude, we make the following contributions:
\begin{itemize*}
\item We introduce \scrl, a novel RL framework specifically designed for training LLM agents in real web environments, enabling iterative reasoning and search, and synthesizing diverse web information to answer open-domain questions.
\item We overcome numerous technical challenges inherent to RL scaling with real-world web search, including API rate limitations, webpage parsing variability, anti-crawling mechanisms, and network latency issues, making this the first successful implementation of reinforcement learning at scale in genuine web environments.
\item We conduct extensive experiments across open-domain tasks, demonstrating significant improvements over prompt-engineered baselines and RAG-based RL approaches.
\item We perform detailed analysis examining emergent behaviors from \scrl's end-to-end RL scaling, finding that the system can formulate plans, cross-validate answers, reflect on its process, and maintain honesty about limitations.
\item We open-source our complete training framework to the research community, fostering transparency and enabling further advancements in deep research systems.
\end{itemize*}

\section{Related Work}

In this section, we review existing approaches to enhance large language models' (LLMs) ability to access external knowledge with search. We categorize these methods into prompt-based and training-based search agents. We also examine the environments in which these methods operate—either local retrieval-augmented generation (RAG) or real-world web search—and position our work in this landscape.

\subsection{Prompt-Based Search Agents}
Many current approaches rely on manually crafted workflows that specify how LLMs should interact with external knowledge sources \cite{wang2024searching}. Recent works such as OpenResearcher~\cite{zheng-etal-2024-openresearcher}, AirRAG~\cite{feng2025airrag}, IterDRAG~\cite{yue2024inference}, Plan*RAG~\cite{verma2025planragefficienttesttimeplanning}, Search-o1 \cite{li2025search} and Open Deep Search \cite{alzubi2025open} have demonstrated significant progress in search capabilities through carefully designed workflows. However, these methods face inherent limitations due to their reliance on human-engineered prompts and interaction patterns, resulting in strict behavior patterns that limit adaptability.

\subsection{Training-Based Search Agents}
Recent developments have moved beyond manually crafted prompts toward training-based approaches that enable more flexible and adaptive search behaviors.

\paragraph{Supervised Fine-Tuning (SFT)}
SFT for RAG have become an enhanced alternative to manual optimization of RAG workflows~\cite{yu2024auto, wang2024corag}. For example, CoRAG \cite{wang2024corag} utilizes Monte Carlo Tree Search (MCTS) to dynamically select the best document blocks under budget constraints. However, it faces limitations including high computational overhead due to MCTS and weak generalization to unknown scenarios due to the dependence on supervised signals.

\paragraph{Reinforcement Learning (RL)}
End-to-end reinforcement learning offers a promising alternative that effectively unlocks LLMs' inherent capabilities. By late 2024, large language models achieved remarkable breakthroughs in reasoning capability enhancement through RL~\cite{guo2025deepseek, openai-o1, team2025kimi}. Recent research has explored applying RL to external knowledge retrieval, with systems such as Search-R1~\cite{jin2025search}, ReSearch~\cite{chen2025research}, and R1-Searcher~\cite{song2025r1} abandoning manually specified cues in favor of models that autonomously develop reasoning during the retrieval process. While OpenAI has acknowledged using RL techniques to enhance their research agent's capabilities, detailed methodologies remain proprietary and undisclosed, creating a significant gap in open research.

\subsection{Training Environments}
Training environments for search agents can be broadly categorized into two types:

\paragraph{Local RAG Environments}
Current mainstream local RAG frameworks \cite{gao2023retrieval, yu2024auto} rely on pre-built fixed knowledge repositories, resulting in three critical issues: information timeliness decay, poor domain adaptability, and storage efficiency bottlenecks. While RAG-based RL approaches like Search-R1~\cite{jin2025search}, ReSearch~\cite{chen2025research}, and R1-Searcher~\cite{song2025r1} have made progress, their experimental validation remains primarily confined to predefined knowledge bases and similarity-based search, restricting the search space and potentially limiting generalizability to real-world applications.

\paragraph{Real-World Web Search Environments}
Web search-based methods \cite{schick2023toolformer, qin2023toolllm} integrate open search engines with LLMs to access real-time information. These approaches face unique challenges including managing search API rate limits, handling network latency, and processing diverse webpage structures. Despite these challenges, real-world environments offer unstructured and heterogeneous information sources that better reflect the complexity of actual research tasks. However, search-based methods requiring external system participation are seldom trained end-to-end, with research often gravitating toward optimization through manually crafted workflows \cite{wang2024searching}.

In contrast to previous work, our approach uniquely combines reinforcement learning with training in genuine web environments. Unlike existing RL methods that primarily focus on static, local text corpora, our method trains agents to interact directly with live search engines. This enables them to handle the inherent variability and complexity of the open web, developing robust capabilities for real-world information retrieval and synthesis. Our approach addresses the limitations of both prompt-based and RAG-confined methods by learning adaptive search strategies through direct interaction with the unpredictable nature of web environments.

\section{Methodology}

\begin{figure}
    \centering
    \includegraphics[width=\linewidth]{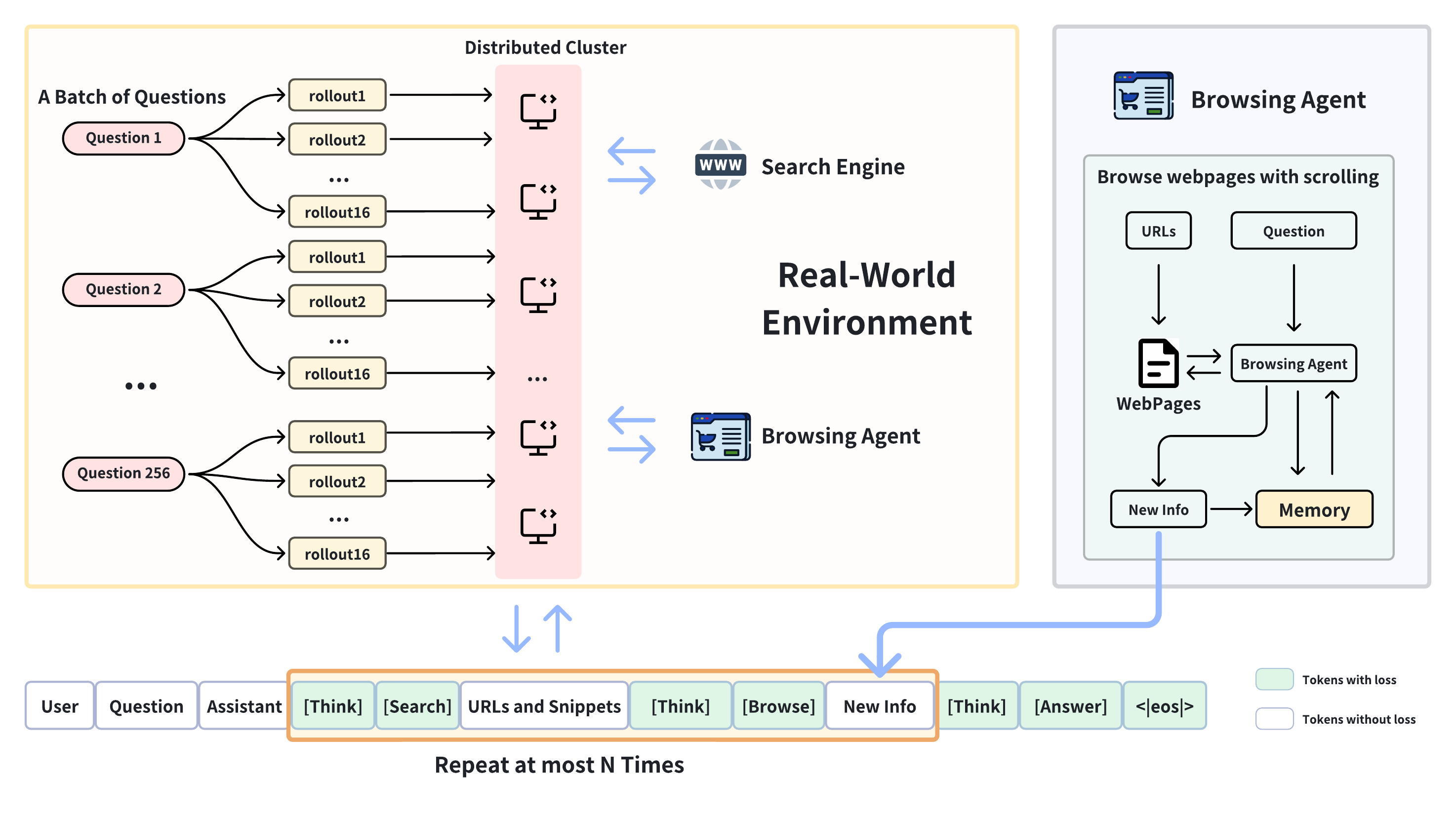}
    \caption{The trajectory of a single sample from a batch of questions processed in parallel by a distributed cluster. Each question undergoes multiple independent rollouts with distinct memory. Upper-left: Displays the batch of questions and their concurrent rollout paths. Upper-right: Shows the browsing agent retrieving web pages via URLs, processing them sequentially to incrementally extract relevant information. Bottom: Details the iterative decision-making steps, from initial query formulation and search to snippet retrieval, further reasoning, browsing, information extraction, and answer generation.}
    \label{fig:training_framework}
\end{figure}

In this section, we describe the methodology used to train an agent capable of solving problems with web search in dynamic real-world environments.

\subsection{Deep Research Trajectory}

In a \scrl's trajectory, it conducts reasoning and tool selection based on the user question and observations iteratively as illustrated in Figure~\ref{fig:training_framework}.

\paragraph{Reasoning}
We restrict \scrl to do reasoning before taking action. Each reasoning process is wrapped in a \texttt{<think>} tag following the setting in DeepSeek-R1~\cite{guo2025deepseek}. 

\paragraph{Web Search Tool} \scrl invokes the web search tool by generating a JSON-formatted request with the tool name \texttt{web\_search} and the search queries as arguments. Search results are returned in a structured format comprising title, URL, and snippet for each webpage. The current implementation employs a fixed top-k (e.g., 10) value for search results retrieval. Future work could explore LLM-driven dynamic parameter optimization for enhanced search efficacy.

\paragraph{Web Browsing Agent}
The web browsing agent provides reliable, question-relevant, and incrementally updated information in to the \scrl system. Specifically, the agent maintains a short-term memory repository for each query. Upon receiving a \texttt{web\_browse} request, it processes the first-page segment of the URL in the request.
Subsequently, the web browsing agent takes two actions based on the query, historical memory, and the newly acquired webpage content: (1) determining whether to continue reading the next URL/segment or stop and (2) appending relevant information to the short-term memory. Once the agent decides to discontinue further browsing, it compiles all newly added information from the short-term memory and returns it to the \scrl system.

\paragraph{Answering}
When the model determines it has sufficient information to answer the question, it generates a final response within \textless answer\textgreater \textless/answer\textgreater
 as the answer to return to the user.

\subsection{Addressing Challenges in Dynamic Real-World Web Environments}
In our open, real-world web setting, several unique challenges arise that necessitate specialized solutions. The following sections detail our strategies for managing these issues effectively.

\paragraph{Challenge I: High-concurrency requests at a single moment}

The implementation of GRPO results in a large number of sampling iterations, leading to a significant volume of search queries and webpage crawling operations (e.g., 4096), causing long delays. To resolve this issue, we created a distributed CPU server cluster with 50 nodes, specifically designed to manage the Tool requests generated during the RL rollout process. Each server is tasked with handling a portion of these requests, processing search results, and crawling webpages based on the URLs identified by the language model for further reading.

\paragraph{Challenge II: Managing Web Crawling and API Limitations}

During the crawling phase, the system frequently encounters anti-crawl measures deployed by web servers, which may return irrelevant content or fail to respond entirely. Similarly, when interfacing with search engines or LLM APIs, restrictions such as provider rate limits (e.g. 200 per second) can arise. To mitigate these issues, we implemented a robust retry mechanism that effectively addresses exceptions encountered during API calls or webpage crawling. In addition, we introduced a caching strategy for search results: if an identical search query is made within a predetermined period (e.g., 7 days), the system retrieves the results from the cache. This approach not only reduces the API call frequency but also helps manage the associated costs, particularly for expensive services like the Google Search API.

\paragraph{Challenge III: Optimizing Information Extraction via a Multi-Agent Approach}

We employ a multi-agent framework wherein a dedicated reading agent is tasked with extracting pertinent information from crawled webpages. Given that many webpages are lengthy and may contain limited relevant content, these pages are partitioned into smaller segments. The reading agent mimics human behavior by processing content sequentially from the first page onward. Under the assumption that if the initial segments of a URL predominantly contain irrelevant information, the webpage is likely unproductive and can be skipped, this method enables more efficient resource allocation and improves overall information extraction accuracy.

\subsection{RL Training Framework}
Our approach utilizes Reinforcement Learning (RL) to train the agent. This section outlines how we employ the RL framework to train the agent and the tools used within it.

\paragraph{GRPO}
In this work, we adopt the \textbf{Group Relative Policy Optimization (GRPO)} algorithm. GRPO optimizes the current policy \(\pi_{\theta}\) by leveraging a reference policy \(\pi_{\theta_{\text{ref}}}\) along with a set of rollouts generated by an existing policy \(\pi_{\theta_{\text{old}}}\). Specifically, given \(G\) rollouts
\[
\tau = \{y_i\}_{i=1}^G \sim \pi_{\theta_{\text{old}}}(\cdot|x)
\]
(with each input \(x \sim D\), where \(D\) is the experience distribution), GRPO estimates the baseline using these trajectories instead of training a separate critic. The current policy is then optimized by maximizing the following objective function:

\begin{equation}
\begin{aligned}
\label{eq:grpo}
  \mathcal{J}(\theta) & = \mathbb{E}_{x \sim \mathcal{D}, \{y_i\}_{i=1}^{G} \sim \pi_{\theta_{\text{old}}}(\cdot|x)} \\
  & \frac{1}{G} \sum_{i=1}^{G} \left[\min \left( \frac{\pi_{\theta}(y_i|x)}{\pi_{\theta_{\text{old}}}(y_i|x)} A_{i}, \text{clip} \left( \frac{\pi_{\theta}(y_i|x)}{\pi_{\theta_{\text{old}}}(y_i|x)}, 1-\epsilon, 1+\epsilon \right) A_{i} \right) - \beta \mathbb{D}_{\text{KL}} \left( \pi_{\theta} || \pi_{\theta_{\text{ref}}} \right)\right],
\end{aligned}
\end{equation}

\paragraph{Masking Observations}
The output of the tool is an observation, not the desired result that the model is expected to produce. Therefore, we apply masking to prevent the observation from being involved in training, allowing only the model's responses to contribute to the training process.

\subsection{Reward}
Rewards play a crucial role during the training process, guiding the agent to continuously improve its performance. This section defines the reward structure and describes how the agent's behavior is rewarded.

We employ F1 score as our primary reward metric due to our utilization of open-domain QA datasets with short-answer ground truth. For future work involving long-form answers, more sophisticated reward mechanisms may be necessary, as noted in the Deep Research system card~\cite{openai2025deep}. The reward is determined by the following conditions:

\[
\text{reward} = 
\begin{cases}
-1 & \text{if format is incorrect} \\
\text{F1 score} & \text{if format is correct}
\end{cases}
\]

\begin{itemize}
  \item \textbf{Format Penalty}: If the format is incorrect (e.g., missing tags or structural errors), the agent receives a penalty of -1.
  \item \textbf{F1 Reward}: If the format is correct, the reward is based on the word-level F1 score, which measures the accuracy of the generated answer compared to the reference answer. A higher F1 score results in a higher reward.
\end{itemize}

\section{Beyond Memorization: Curating Search-Dependent Training Data}
\label{sec:training_data}
\subsection{Leveraging Open Domain QA Data}
Despite the growing interest in deep research capabilities for LLM agents, there currently exists no open-source training dataset specifically designed for this purpose. To address this gap, we leverage existing open-domain question-answering datasets, which contain single-hop to multi-hop questions that inherently require online search to find accurate answers. 

Our training corpus comprises a diverse collection of QA datasets that require varying degrees of retrieval complexity. Specifically, we utilize NaturalQuestions (NQ)~\cite{kwiatkowski-etal-2019-natural} and TriviaQA (TQ)~\cite{joshi-etal-2017-triviaqa} for single-hop scenarios, where answers can typically be found within a single web document. For more complex multi-hop scenarios, which require integrating information across multiple sources, we incorporate examples from HotpotQA~\cite{yang2018hotpotqa} and 2WikiMultiHopQA (2Wiki)~\cite{xanh2020_2wikimultihop}, both of which were specifically designed to evaluate multi-step reasoning capabilities.

\subsection{The Issue of Data Contamination}
For training models that genuinely learn to leverage web search tools—rather than simply recalling memorized information—it is critical to address the problem of data contamination. Large language models have been pretrained on vast internet corpora, which likely include many of the QA pairs in standard benchmarks. Without proper contamination detection, the model might appear to successfully complete research tasks while actually using its parametric knowledge, defeating the purpose of learning web search strategies.

This contamination issue is particularly problematic in the context of our work, as it could lead to:
\begin{itemize}
\item Models that falsely appear to benefit from web search when actually using memorized knowledge.
\item Failure to develop genuine search strategies when deployed on truly novel questions.
\item Inability to generalize to real-world research scenarios where answers cannot be found in the model's training data.
\end{itemize}

\subsection{Data Cleaning and Contamination Detection}
To ensure the integrity of our training process, we implemented a comprehensive two-stage filtering methodology:

\paragraph{Low-Quality Question Filtering} We exclude questions that could yield unreliable or problematic search results. Specifically, we eliminate: 1) Time-sensitive questions (e.g., "Who is the current CEO of Apple?"); 2) Highly subjective queries (e.g., "What is the best smartphone?"); and 3) Potentially harmful or policy-violating content. This filtering was implemented using DeepSeek-R1~\cite{guo2025deepseek} with a carefully designed evaluation prompt to systematically identify and mark problematic questions.

\paragraph{Contamination Detection} To ensure the model genuinely learns to use search tools rather than memorizing answers, we employed a robust contamination detection procedure. For each candidate question, we randomly sample 10 responses from the base model we will use in training, and check if any response contains the ground truth answer (i.e., pass@10). Questions where the model demonstrated prior knowledge (by producing the correct answer without search) were excluded from the training set. This contamination screening is critical for preventing the model from developing a false reliance on parametric knowledge when search-based knowledge is required.

The prompts used for data cleaning and contamination detection are listed in Appendix~\ref{appendix:prompt_quality_control}. After applying these quality control measures, we constructed a final training dataset of 80,000 examples with a distribution ratio of 1:1:3:3 for NQ:TQ:HotpotQA:2Wiki. This proportion deliberately emphasizes multi-hop scenarios (75\% of examples), as these better reflect the complex information-seeking behaviors required for deep research questions.

\section{Experiments}

\subsection{Experimental Setups}

\subsubsection{Model and Hyperparameters}

We adopt Qwen2.5-7B-Instruct\footnote{https://huggingface.co/Qwen/Qwen2.5-7B-Instruct} \cite{qwen2025qwen25technicalreport} as the backbone model for our training pipeline. The training is conducted using the verl framework\footnote{https://github.com/volcengine/verl}. At each training step, we sample 256 prompts, and sample 16 rollouts for each prompt. Each rollout consists of up to 10 tool calls followed by a final answer step. The training is performed with a mini-batch size of 4,096, which means one rollout stage will backprop for one time.

\subsection{Evaluation and Results}

\subsubsection{Benchmarks}

To thoroughly evaluate model performance across both in-domain (ID) and out-of-domain (OOD) settings, we construct a diverse benchmark suite spanning a range of open-domain QA challenges. For in-domain evaluation, we include the dev sets of NQ \cite{kwiatkowski-etal-2019-natural}, TQ \cite{joshi-etal-2017-triviaqa}, HotpotQA \cite{yang2018hotpotqa}, and 2Wiki \cite{xanh2020_2wikimultihop} as mentioned in Section~\ref{sec:training_data}.

For out-of-domain evaluation, we introduce three datasets that differ significantly in question style and information distribution: MuSiQue \cite{trivedi2021musique}, Bamboogle \cite{press2022measuring}, and PopQA \cite{mallen2023llm_memorization}. These datasets test the model’s generalization ability beyond the training domain.

To ensure a fair and balanced evaluation, we randomly sample 512 examples from the development sets of NQ, TQ, HotpotQA, 2Wiki, MuSiQue, and PopQA as well as all 125 samples from Bamboogle's development set. This sampling strategy allows us to assess model robustness across a broad range of topics and reasoning requirements.

\subsubsection{Baselines}

To evaluate the effectiveness of \scrl, we compare it against the following baseline methods:
\begin{itemize}
    \item \textbf{CoT Only}: This baseline employs Chain-of-Thought (CoT) \cite{} reasoning to generate answers without access to any external reference context.
    \item \textbf{RAG}: This approach combines Chain-of-Thought reasoning with retrieved reference context to guide the answer generation process.
    \item \textbf{Search-o1}: A multi-step reasoning baseline in which the model generates search queries or intermediate answers. For each query, the context is limited to a snippet retrieved by a retriever, rather than full documents.\footnote{To ensure consistency with other results, we reimplemented search-o1 using our own prompt.}

    \item \textbf{Search-o1 + Web Search}: In contrast to Search-o1, this setting lets the model access the open web. It can send real-time search queries through APIs like Serper and visit URLs to browse webpages. This supports richer, more dynamic information gathering and forms the basis of deep research.

    \item \textbf{Search-r1}: A reinforcement learning method for question answering. During the training and inference stages, it searches Wikipedia information with the help of a retriever. There are two versions: Search-r1-base and Search-r1-instruct, where the initial actor model is either the base model or the instruct model.

    \item \textbf{R1-Searcher}: Unlike Search-r1, when given a search query, it appends \texttt{"site:en.wikipedia.org"} to the query, search it via Bing, and summarizes the first three pages of the search results. \scrl differs from this approach in three key aspects: (1) \scrl is also trained with real-world environment; (2) \scrl does not restrict the search space to a specific domain, such as Wikipedia; and (3) Our method allows the model to autonomously select URLs rather than compulsorily summarizing the top three search results.
\end{itemize}

\subsubsection{Evaluation Metrics}

\paragraph{Rule-based Metrics}
We evaluate model performance using F1 score aligning with the reward for training. Both ground-truth and predicted answers are normalized by converting to lowercase and removing all punctuation before computing the metrics.

\paragraph{Model-based Evaluation}
Rule-based evaluation doesn't suit long-form responses, so we adopt a model-based evaluation (MBE) approach using LLM-as-a-Judge \cite{DBLP:conf/nips/ZhengC00WZL0LXZ23}. Specifically, we prompt GPT-4o-mini \cite{hurst2024gpt} to assess the model's answer against the question and ground truth answer, and label it as either "correct" or "incorrect." The MBE score is then computed as the accuracy of these judgments.\cite{DBLP:conf/nips/ZhengC00WZL0LXZ23} The full prompt is provided in Appendix~\ref{emb prompt}.

\begin{table}
    \centering
    \scriptsize
    \resizebox{\textwidth}{!}{
    \begin{tabular}{lc|cc|cc|cc|cc}
\toprule
\multirow{2}{*}{\textbf{Method}} & \multirow{2}{*}{\textbf{\shortstack{Inference \\ Environment}}} & \multicolumn{2}{c}{\textbf{NQ}} & \multicolumn{2}{c}{\textbf{TQ}} & \multicolumn{2}{c}{\textbf{HotpotQA}} & \multicolumn{2}{c}{\textbf{2Wiki}} \\
\cline{3-10}
& & F1 & MBE & F1 & MBE & F1 & MBE & F1 & MBE \\
\hline
\textit{\textbf{Prompt Based}} \\
CoT & Local RAG & 19.8 & 32.0& 45.6 & 48.2  & 24.4 & 27.9  & 26.4 & 27.3 \\
CoT + RAG  & Local RAG & 42.0 & 59.6 & 68.9 & 75.8 & 37.1 & 43.8 &  24.4 & 24.8  \\
Search-o1*  & Local RAG  & 34.5 & 57.4 & 52.6 & 61.1 & 31.6 & 40.8 &  28.6 & 32.8  \\
Search-o1  & Web Search & 32.4 & 55.1 & 58.9 & 69.5 &  33.0 & 42.4 &  30.9 & 37.7 \\
\hline
\textit{\textbf{Training Based}} \\
Search-r1-base  & Local RAG & \textbf{45.4} & 60.0 & 71.9 & 76.2 & \textbf{55.9} & 63.0 & 44.6 & 47.9 \\
Search-r1-instruct  & Local RAG  & 33.1 & 49.6  & 44.7 & 49.2 & 45.7 & 52.5 & 43.4 & 48.8 \\
R1-Searcher   & Web Search  & 35.4 & 52.3  & 73.1 & 79.1  & 44.8 & 53.1 & 59.4 & 65.8  \\
\rowcolor[rgb]{0.8,0.9,1.0}  \scrl  & Web Search & 39.6 & \textbf{61.9}  & \textbf{78.4} & \textbf{85.0}  & 52.8 & \textbf{64.3}  & \textbf{59.7} & \textbf{66.6} \\
\bottomrule
    \end{tabular}
    }
    \caption{In-domain results on four datasets (NQ, TQ, HotpotQA, 2Wiki), evaluated by F1 and MBE metrics. DeepResearcher outperforms all baseline methods in MBE and shows competitive performance in F1, particularly excelling on TQ and 2Wiki. It is worth noting that Search-r1-base was trained and evaluated in a local RAG environment with direct access to the relevant Wikipedia corpus, while DeepResearcher must navigate the entire Internet to find information, achieving excellent results despite facing a more realistic and challenging scenario.}
    \label{tab:main_results_in_domain}
\end{table}

\subsubsection{Main Results}

Table~\ref{tab:main_results_in_domain} and Table~\ref{tab:main_results_out_of_domain} present the main results of \scrl and the baselines in-domain and out-of-domain, respectively. From these results, we draw the following observations:

\paragraph{\scrl outperforms the baselines within training domains.} As shown in Table~\ref{tab:main_results_in_domain}, \scrl achieves the highest performance across the four datasets when measured by the more reliable MBE metric, outperforming baselines by a substantial margin on TQ and 2Wiki. While Search-r1-base shows comparable MBE results on NQ and HotpotQA, it's important to note that Search-r1-base was specifically trained and evaluated using a local RAG system with direct access to the relevant Wikipedia corpus. In contrast, \scrl must navigate the entire Internet to find relevant information, representing a more realistic and significantly more challenging scenario even though the answers ultimately come from Wikipedia.

\paragraph{\scrl demonstrates exceptional generalization to novel domains.} As revealed in Table~\ref{tab:main_results_out_of_domain}, \scrl consistently outperforms all other baselines across three OOD datasets. This indicates that the model successfully learns generalizable skills for reasoning, searching, and synthesizing information from different sources through RL scaling, rather than merely adapting to specific training distributions.

\paragraph{Importance of Real-World Environment in Training} Questions in Bamboogle specifically require knowledge beyond Wikipedia's coverage. Consequently, \scrl significantly outperforms local RAG-based methods on this benchmark. Furthermore, even when we enable R1-Searcher (which was trained using local RAG) to search the real-world web, it still performs substantially worse than \scrl. These results demonstrate the critical advantage of using real-world environments during RL scaling training, as this exposure develops robust information retrieval and synthesis capabilities that cannot be achieved in controlled, static environments.

\begin{table}
    \centering
    \small
    \resizebox{0.8\textwidth}{!}{ 
    \begin{tabular}{lc|cc|cc|cc}
\toprule
\multirow{2}{*}{\textbf{Method}} & \multirow{2}{*}{\textbf{\shortstack{Inference \\ Environment}}} & \multicolumn{2}{c}{\textbf{Musique}} & \multicolumn{2}{c}{ \cellcolor[rgb]{1, 0.9, 0.8}\textbf{Bamboogle}} & \multicolumn{2}{c}{\textbf{PopQA}}  \\
\cline{3-8}
 &  & F1 & MBE & \cellcolor[rgb]{1, 0.9, 0.8}F1 & \cellcolor[rgb]{1, 0.9, 0.8}MBE  & F1 & MBE \\
\hline
\textit{\textbf{Prompt Based}} & & & & \cellcolor[rgb]{1, 0.9, 0.8} &\cellcolor[rgb]{1, 0.9, 0.8} & & \\
CoT  & Local RAG & 8.5 & 7.4  & \cellcolor[rgb]{1, 0.9, 0.8}22.1 & \cellcolor[rgb]{1, 0.9, 0.8}21.6  & 17.0 & 15.0   \\
CoT + RAG  & Local RAG& 10.0 & 10.0  & \cellcolor[rgb]{1, 0.9, 0.8}25.4 & \cellcolor[rgb]{1, 0.9, 0.8}27.2  & 46.9 & 48.8 \\
Search-o1*  & Local RAG & 16.8 & 21.3  & \cellcolor[rgb]{1, 0.9, 0.8}35.8 & \cellcolor[rgb]{1, 0.9, 0.8}38.4 & 36.9 & 42.4   \\
Search-o1  & Web Search  & 14.7 & 19.7  & \cellcolor[rgb]{1, 0.9, 0.8}46.6 & \cellcolor[rgb]{1, 0.9, 0.8}53.6  & 38.3 & 43.4   \\
\hline
\textit{\textbf{Training Based}} & & & & \cellcolor[rgb]{1, 0.9, 0.8} &\cellcolor[rgb]{1, 0.9, 0.8} & & \\
Search-r1-base  & Local RAG  & 26.7 & 27.5 & \cellcolor[rgb]{1, 0.9, 0.8}56.5 & \cellcolor[rgb]{1, 0.9, 0.8}57.6 & 43.2 &  47.0 \\
Search-r1-instruct  & Local RAG  & 26.5 & 28.3  & \cellcolor[rgb]{1, 0.9, 0.8}45.0 & \cellcolor[rgb]{1, 0.9, 0.8}47.2  & 43.0 &  44.5  \\
R1-Searcher & Web Search  & 22.8 & 25.6  & \cellcolor[rgb]{1, 0.9, 0.8}64.8 & \cellcolor[rgb]{1, 0.9, 0.8}65.6  & 42.7 & 43.4   \\
\rowcolor[rgb]{0.8,0.9,1.0} \scrl & Web Search & \textbf{27.1} & \textbf{29.3} & \cellcolor[rgb]{0.9, 0.9, 0.9}\textbf{71.0} & \cellcolor[rgb]{0.9, 0.9, 0.9}\textbf{72.8} & \textbf{48.5} & \textbf{52.7}  \\
\bottomrule
    \end{tabular}
    }
    \caption{This table shows the performance of different methods on three out-of-domain datasets (Musique, Bamboogle, PopQA), evaluated by F1 and MBE metrics. \scrl leads in both F1 and MBE on all datasets, demonstrating strong generalization capabilities compared to other methods. Notably, unlike the other datasets, Bamboogle’s corpus is not entirely derived from Wikipedia pages.}
    \label{tab:main_results_out_of_domain}
\end{table}

\section{Analysis}

\subsection{Training Dynamics}
\begin{figure}[htbp]
  \centering
  \begin{tabular}{ccc}
    \includegraphics[width=0.30\textwidth]{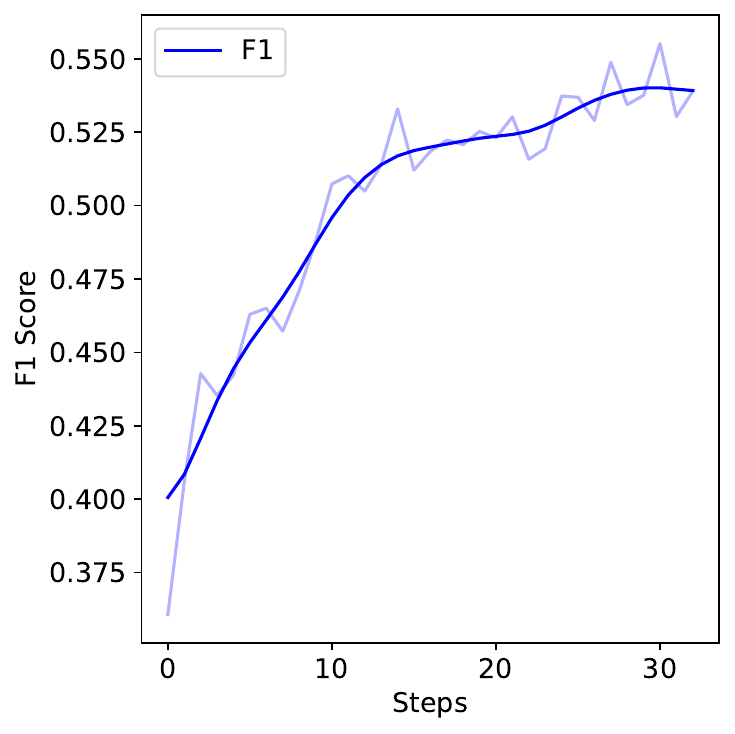} &
    \includegraphics[width=0.30\textwidth]{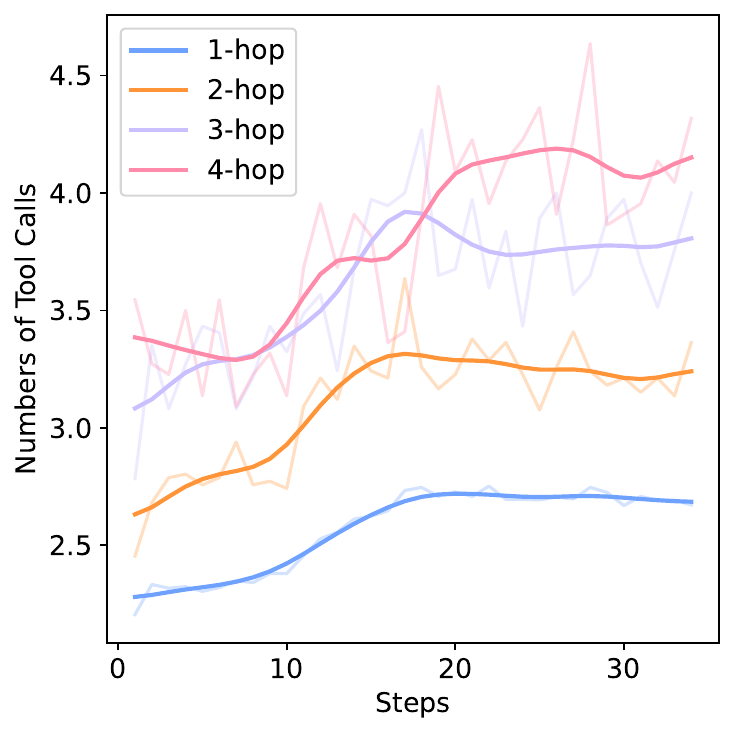} &
    \includegraphics[width=0.30\textwidth]{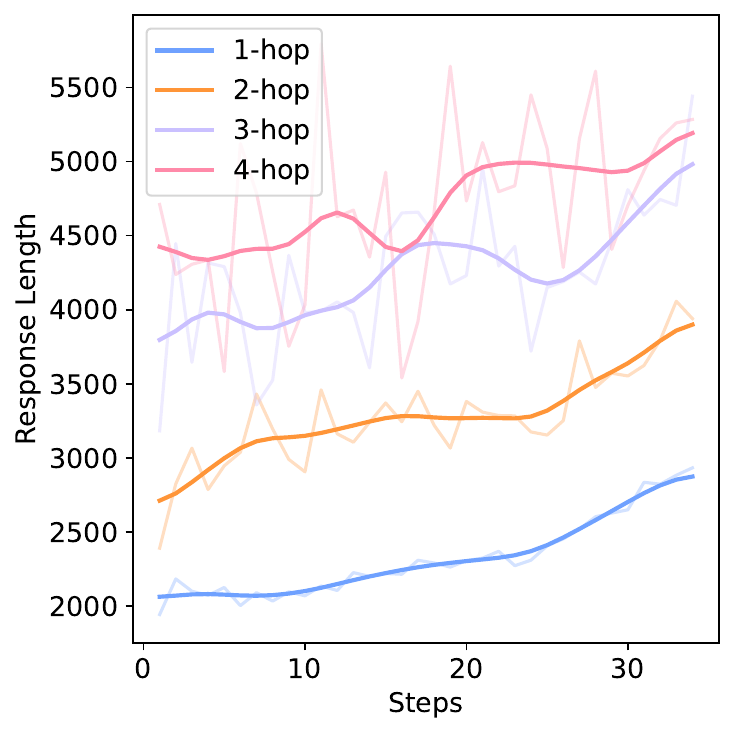} \\
    \;\;\;\;\;\;\;\;(a) & \;\;\;\;\;\;(b) & \;\;\;\;\;\;\;(c) 
  \end{tabular}
  \caption{Training dynamics of F1, turns and response length. In this figure, we find the performance gradually scaling with reinforcement learning. The numbers of tool calls and responses also increase.}
  \label{fig:metrics}
\end{figure}

\begin{itemize}
\item \textbf{Performance gradually scaling with reinforcement learning:} Figure \ref{fig:metrics} (a) present the evaluation of F1 scores, across different training steps.  The F1 score 0.375, and gradually increases to around 0.55 demonstrating a consistent upward trend. This result indicates the progressive improvement of the model's performance in reinforcement learning.

\item \textbf{Training leads to increased reasoning steps in hard question:} Figure \ref{fig:metrics} (b) illustrates the average number of turns required for different reasoning hops. The general trend indicates that as the training progresses, the required number of tool calls also increases across different difficulty levels. Unlike the other three settings, the 4-hop setting continues to exhibit an increasing trend even after 34 steps. This suggests that the model is still learning to retrieve more information when dealing with more difficult questions.

\item \textbf{Continuous learning makes long response without saturation:} Figure \ref{fig:metrics} (c) presents the length of responses for different reasoning hops. The response lengths also increase with reasoning complexity. However, all four settings show a sustained upward trend, indicating that the model continues to expand its reasoning processes during training. This further supports the idea that the model adapts to increasingly complex queries by generating more detailed outputs like double-check, refinement, planning, etc.
\end{itemize}

\begin{figure}
    \centering
    \includegraphics[width=\linewidth]{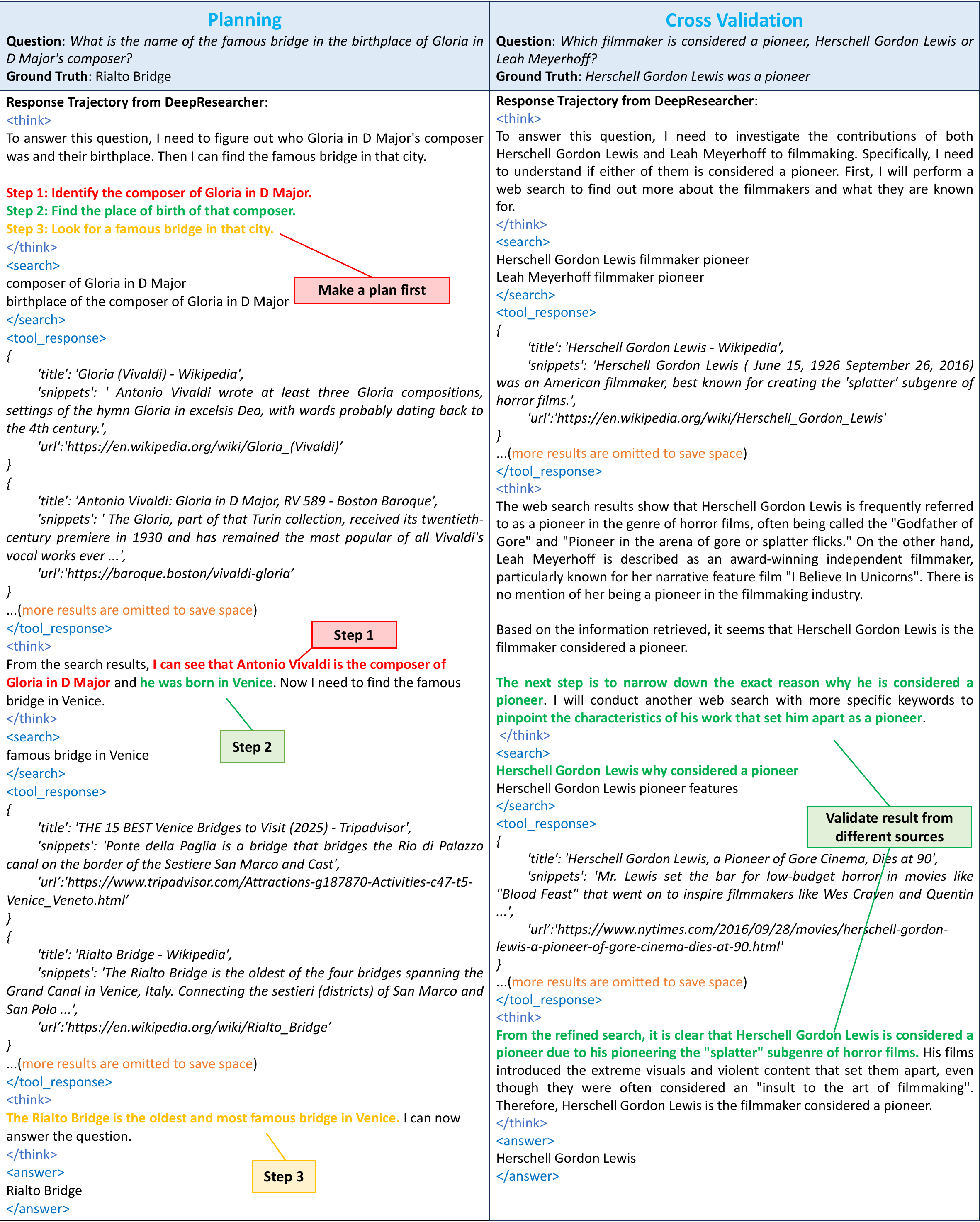}
    \caption{Planning and Cross Validation Behaviors. \textbf{Left}: \scrl is capable of making plans and dynamically adjusting it throughout the reasoning process. Notably, the model can merge steps when appropriate. \textbf{Right}: \scrl identifies the correct answer during its first tool call. However, rather than immediately committing to this result, it proceeds to verify its accuracy through subsequent steps. This rigorous behavior enhances the reliability of the model's responses,  ensuring greater robustness in its final answers. Note: In the actual model, the tool call and tool response format is a JSON string, rather than a tag. However, for clarity and ease of presentation, we have used tags in the figure to represent the output.}
    \label{fig:case_study_}
\end{figure}

\begin{figure}
    \centering
    \includegraphics[width=\linewidth]{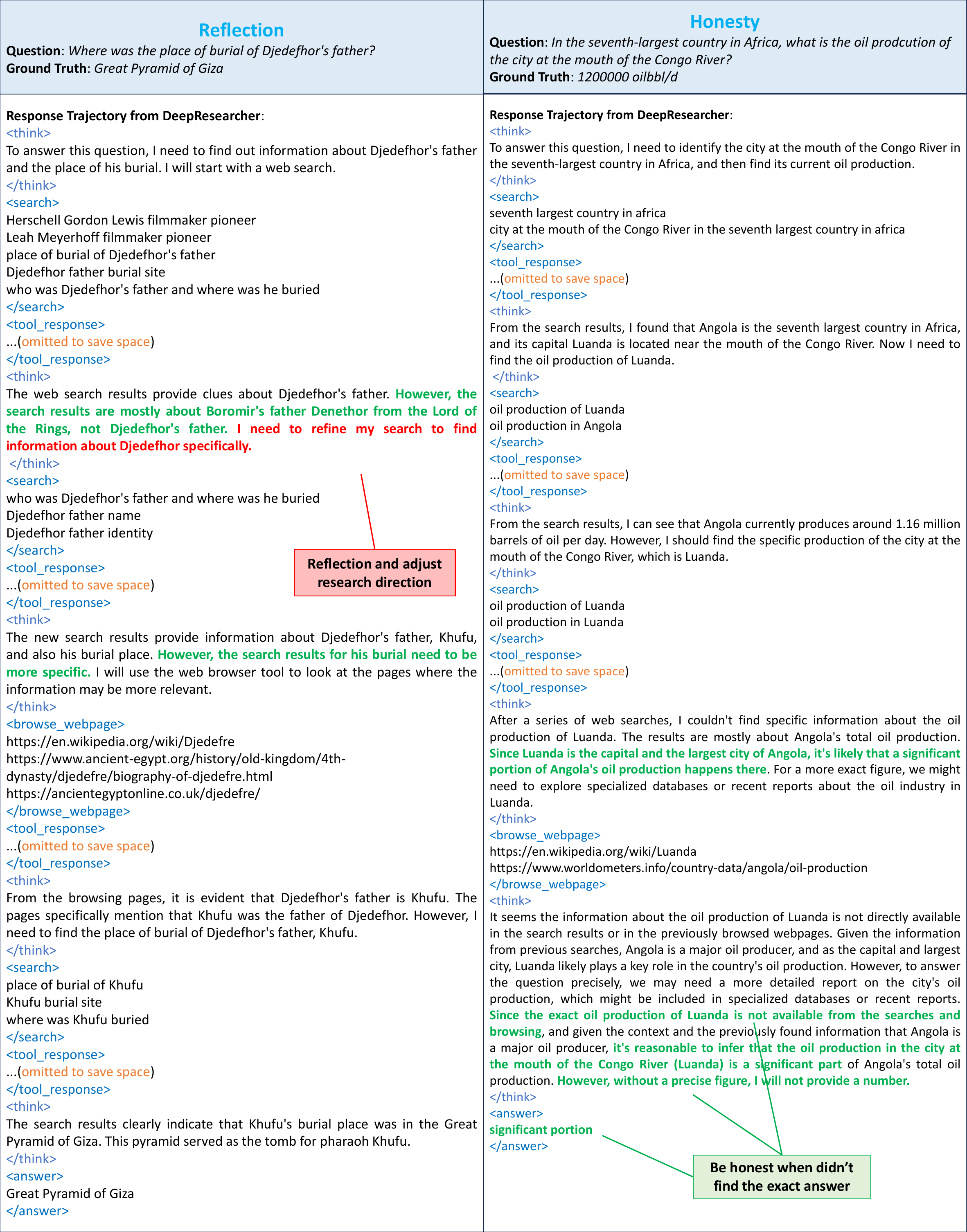}
    \caption{Reflection and Honesty Behavior. The search and browse are 2 apis in json format in the real inference stage. Left:  When the retrieved information does not fully align with the question, \scrl recognizes this discrepancy based on environmental feedback and refines its search query in subsequent tool calls. This proves its reflection ability. Right: \scrl is capable of recognizing when it has not found the correct answer and appropriately declines to provide a response to be honesty. Note: In the actual model, the tool call and tool response format is a JSON string, rather than a tag. However, for clarity and ease of presentation, we have used tags in the figure to represent the output.}
    \label{fig:reflection-label}
\end{figure}

\subsection{Case Study}

Figures~\ref{fig:planning-label} and \ref{fig:reflection-label} present four cases illustrating the model's behavior after reinforcement learning. From these examples, we identify several key behavioral patterns:

\begin{itemize}
\item \textbf{Behavior I: Planning when addressing multi-hop questions:} As demonstrated on the left side of Figure~\ref{fig:planning-label}, \scrl is capable of making plans and dynamically adjusting it throughout the reasoning process. Notably, the model can merge steps when appropriate, indicating that planning abilities can emerge naturally without the necessity of SFT on explicit planning data \cite{yue2024dots}.
\item \textbf{Behavior II: Cross-validation before finalizing its answers:} As observed on the right side of Figure~\ref{fig:planning-label}, \scrl identifies the correct answer during its first tool call. However, rather than immediately committing to this result, it proceeds to verify its accuracy through subsequent steps. This cautious approach enhances the reliability of the model's responses,  ensuring greater robustness in its final predictions.
\item \textbf{Behavior III: Reflection when observations deviate from expectations:} The left side of Figure~\ref{fig:reflection-label} illustrates the model's ability to reflect on its search process. When the retrieved information does not fully align with the question, \scrl recognizes this discrepancy based on environmental feedback and refines its search query in subsequent tool calls. This reflective capability is essential for preventing the model from getting stuck \cite{fu2025agentrefine} in reasoning, enabling it to enhance overall problem-solving efficiency.
\item \textbf{Behavior IV: Honesty by acknowledging its limitations:} A reliable model should minimize hallucinations and provide honest responses when it lacks the necessary knowledge \cite{yang2024alignment}. We observe that \scrl is capable of recognizing when it has not found the correct answer and appropriately declines to provide a response. This behavior is beneficial, however, current question-answering evaluation metrics do not yet account for this aspect of model reliability.
\end{itemize}

\section{Conclusion}
In conclusion, we presents \scrl, a groundbreaking approach for scaling reinforcement learning in LLMs to operate effectively in real-world web search environments. Unlike existing methods that rely on static knowledge bases or controlled retrieval settings, \scrl trains agents to interact directly with live search engines, allowing them to navigate the inherent complexity and variability of the open web. This direct engagement with dynamic search environments leads to substantial improvements in task completion and research capabilities compared to both prompt-engineered and RAG-based RL agents.

By adopting an end-to-end training framework, \scrl moves beyond human-engineered workflows, empowering the agent to autonomously develop problem-solving strategies. Our approach not only addresses the unique challenges of real-world web search, such as network latency and anti-crawling mechanisms, but also provides a robust multi-agent architecture that enhances the agent’s ability to collect diverse information from the web. The resulting system demonstrates notable cognitive behaviors such as planning, cross-validation, reflection, and maintaining honesty, which are crucial for autonomous agents conducting deep research.

The success of \scrl marks a significant milestone in the evolution of LLM agents, showing that scaling reinforcement learning in real-world environments can unlock substantial improvements in research performance. This approach offers a promising path forward for building more adaptive, intelligent systems capable of solving complex, open-domain problems that are relevant to real-world applications.
\bibliographystyle{acl_natbib}
\bibliography{related}

\begin{thebibliography}{41}
\expandafter\ifx\csname natexlab\endcsname\relax\def\natexlab#1{#1}\fi

\bibitem[{Alzubi et~al.(2025)Alzubi, Brooks, Chiniya, Contente, von Gerlach, Irwin, Jiang, Kaz, Nguyen, Oh et~al.}]{alzubi2025open}
Salaheddin Alzubi, Creston Brooks, Purva Chiniya, Edoardo Contente, Chiara von Gerlach, Lucas Irwin, Yihan Jiang, Arda Kaz, Windsor Nguyen, Sewoong Oh, et~al. 2025.
\newblock Open deep search: Democratizing search with open-source reasoning agents.
\newblock \emph{arXiv preprint arXiv:2503.20201}.

\bibitem[{{Anthropic}(2024)}]{anthropic2024building}
{Anthropic}. 2024.
\newblock \href {https://www.anthropic.com/engineering/building-effective-agents} {Building effective agents}.

\bibitem[{{CAMEL-AI.org}(2025)}]{owl2025}
{CAMEL-AI.org}. 2025.
\newblock Owl: Optimized workforce learning for general multi-agent assistance in real-world task automation.
\newblock \url{https://github.com/camel-ai/owl}.
\newblock Accessed: 2025-03-07.

\bibitem[{Chen et~al.(2025)Chen, Li, Sun, Zhou, Zhu, Wang, Pan, Zhang, Chen, Yang, Zhou, and Chen}]{chen2025research}
Mingyang Chen, Tianpeng Li, Haoze Sun, Yijie Zhou, Chenzheng Zhu, Haofen Wang, Jeff~Z. Pan, Wen Zhang, Huajun Chen, Fan Yang, Zenan Zhou, and Weipeng Chen. 2025.
\newblock \href {http://arxiv.org/abs/2503.19470} {Research: Learning to reason with search for llms via reinforcement learning}.

\bibitem[{Feng et~al.(2025)Feng, Hao, Zhang, Song, and Wang}]{feng2025airrag}
Wenfeng Feng, Chuzhan Hao, Yuewei Zhang, Jingyi Song, and Hao Wang. 2025.
\newblock Airrag: Activating intrinsic reasoning for retrieval augmented generation via tree-based search.
\newblock \emph{arXiv preprint arXiv:2501.10053}.

\bibitem[{Fu et~al.(2025)Fu, He, Wang, Hong, Gongque, Zeng, Wang, Wang, Cai, and Xu}]{fu2025agentrefine}
Dayuan Fu, Keqing He, Yejie Wang, Wentao Hong, Zhuoma Gongque, Weihao Zeng, Wei Wang, Jingang Wang, Xunliang Cai, and Weiran Xu. 2025.
\newblock Agentrefine: Enhancing agent generalization through refinement tuning.
\newblock \emph{arXiv preprint arXiv:2501.01702}.

\bibitem[{Gao et~al.(2023)Gao, Xiong, Gao, Jia, Pan, Bi, Dai, Sun, Wang, and Wang}]{gao2023retrieval}
Yunfan Gao, Yun Xiong, Xinyu Gao, Kangxiang Jia, Jinliu Pan, Yuxi Bi, Yi~Dai, Jiawei Sun, Haofen Wang, and Haofen Wang. 2023.
\newblock Retrieval-augmented generation for large language models: A survey.
\newblock \emph{arXiv preprint arXiv:2312.10997}, 2.

\bibitem[{{Google}(2024)}]{google2024gemini}
{Google}. 2024.
\newblock \href {https://blog.google/products/gemini/google-gemini-deep-research/} {Gemini deep research}.

\bibitem[{Guo et~al.(2025)Guo, Yang, Zhang, Song, Zhang, Xu, Zhu, Ma, Wang, Bi et~al.}]{guo2025deepseek}
Daya Guo, Dejian Yang, Haowei Zhang, Junxiao Song, Ruoyu Zhang, Runxin Xu, Qihao Zhu, Shirong Ma, Peiyi Wang, Xiao Bi, et~al. 2025.
\newblock Deepseek-r1: Incentivizing reasoning capability in llms via reinforcement learning.
\newblock \emph{arXiv preprint arXiv:2501.12948}.

\bibitem[{Ho et~al.(2020)Ho, Duong~Nguyen, Sugawara, and Aizawa}]{xanh2020_2wikimultihop}
Xanh Ho, Anh-Khoa Duong~Nguyen, Saku Sugawara, and Akiko Aizawa. 2020.
\newblock \href {https://www.aclweb.org/anthology/2020.coling-main.580} {Constructing a multi-hop {QA} dataset for comprehensive evaluation of reasoning steps}.
\newblock In \emph{Proceedings of the 28th International Conference on Computational Linguistics}, pages 6609--6625, Barcelona, Spain (Online). International Committee on Computational Linguistics.

\bibitem[{Hong et~al.(2024)Hong, Zhuge, Chen, Zheng, Cheng, Wang, Zhang, Wang, Yau, Lin, Zhou, Ran, Xiao, Wu, and Schmidhuber}]{hong2024metagpt}
Sirui Hong, Mingchen Zhuge, Jonathan Chen, Xiawu Zheng, Yuheng Cheng, Jinlin Wang, Ceyao Zhang, Zili Wang, Steven Ka~Shing Yau, Zijuan Lin, Liyang Zhou, Chenyu Ran, Lingfeng Xiao, Chenglin Wu, and J{\"u}rgen Schmidhuber. 2024.
\newblock \href {https://openreview.net/forum?id=VtmBAGCN7o} {Meta{GPT}: Meta programming for a multi-agent collaborative framework}.
\newblock In \emph{The Twelfth International Conference on Learning Representations}.

\bibitem[{Hurst et~al.(2024)Hurst, Lerer, Goucher, Perelman, Ramesh, Clark, Ostrow, Welihinda, Hayes, Radford et~al.}]{hurst2024gpt}
Aaron Hurst, Adam Lerer, Adam~P Goucher, Adam Perelman, Aditya Ramesh, Aidan Clark, AJ~Ostrow, Akila Welihinda, Alan Hayes, Alec Radford, et~al. 2024.
\newblock Gpt-4o system card.
\newblock \emph{arXiv preprint arXiv:2410.21276}.

\bibitem[{Jin et~al.(2025)Jin, Zeng, Yue, Wang, Zamani, and Han}]{jin2025search}
Bowen Jin, Hansi Zeng, Zhenrui Yue, Dong Wang, Hamed Zamani, and Jiawei Han. 2025.
\newblock Search-r1: Training llms to reason and leverage search engines with reinforcement learning.
\newblock \emph{arXiv preprint arXiv:2503.09516}.

\bibitem[{Joshi et~al.(2017)Joshi, Choi, Weld, and Zettlemoyer}]{joshi-etal-2017-triviaqa}
Mandar Joshi, Eunsol Choi, Daniel Weld, and Luke Zettlemoyer. 2017.
\newblock \href {https://doi.org/10.18653/v1/P17-1147} {{T}rivia{QA}: A large scale distantly supervised challenge dataset for reading comprehension}.
\newblock In \emph{Proceedings of the 55th Annual Meeting of the Association for Computational Linguistics (Volume 1: Long Papers)}, pages 1601--1611, Vancouver, Canada. Association for Computational Linguistics.

\bibitem[{Kwiatkowski et~al.(2019)Kwiatkowski, Palomaki, Redfield, Collins, Parikh, Alberti, Epstein, Polosukhin, Devlin, Lee, Toutanova, Jones, Kelcey, Chang, Dai, Uszkoreit, Le, and Petrov}]{kwiatkowski-etal-2019-natural}
Tom Kwiatkowski, Jennimaria Palomaki, Olivia Redfield, Michael Collins, Ankur Parikh, Chris Alberti, Danielle Epstein, Illia Polosukhin, Jacob Devlin, Kenton Lee, Kristina Toutanova, Llion Jones, Matthew Kelcey, Ming-Wei Chang, Andrew~M. Dai, Jakob Uszkoreit, Quoc Le, and Slav Petrov. 2019.
\newblock \href {https://doi.org/10.1162/tacl_a_00276} {Natural questions: A benchmark for question answering research}.
\newblock \emph{Transactions of the Association for Computational Linguistics}, 7:452--466.

\bibitem[{Li et~al.(2025{\natexlab{a}})Li, Dong, Jin, Zhang, Zhou, Zhu, Zhang, and Dou}]{li2025search}
Xiaoxi Li, Guanting Dong, Jiajie Jin, Yuyao Zhang, Yujia Zhou, Yutao Zhu, Peitian Zhang, and Zhicheng Dou. 2025{\natexlab{a}}.
\newblock Search-o1: Agentic search-enhanced large reasoning models.
\newblock \emph{arXiv preprint arXiv:2501.05366}.

\bibitem[{Li et~al.(2025{\natexlab{b}})Li, Zou, and Liu}]{li2025limr}
Xuefeng Li, Haoyang Zou, and Pengfei Liu. 2025{\natexlab{b}}.
\newblock \href {http://arxiv.org/abs/2502.11886} {Limr: Less is more for rl scaling}.

\bibitem[{Li et~al.(2025{\natexlab{c}})Li, Zou, and Liu}]{li2025torlscalingtoolintegratedrl}
Xuefeng Li, Haoyang Zou, and Pengfei Liu. 2025{\natexlab{c}}.
\newblock \href {http://arxiv.org/abs/2503.23383} {Torl: Scaling tool-integrated rl}.

\bibitem[{Liang et~al.(2025)Liang, Xiang, Yu, Zhang, and Hong}]{openmanus2025}
Xinbin Liang, Jinyu Xiang, Zhaoyang Yu, Jiayi Zhang, and Sirui Hong. 2025.
\newblock Openmanus: An open-source framework for building general ai agents.
\newblock \url{https://github.com/mannaandpoem/OpenManus}.

\bibitem[{Mallen et~al.(2022)Mallen, Asai, Zhong, Das, Hajishirzi, and Khashabi}]{mallen2023llm_memorization}
Alex Mallen, Akari Asai, Victor Zhong, Rajarshi Das, Hannaneh Hajishirzi, and Daniel Khashabi. 2022.
\newblock When not to trust language models: Investigating effectiveness and limitations of parametric and non-parametric memories.
\newblock \emph{arXiv preprint}.

\bibitem[{OpenAI(2024)}]{openai-o1}
OpenAI. 2024.
\newblock \href {https://openai.com/index/learning-to-reason-with-llms/} {Learning to reason with llms, september 2024}.

\bibitem[{{OpenAI}(2025)}]{openai2025deep}
{OpenAI}. 2025.
\newblock \href {https://cdn.openai.com/deep-research-system-card.pdf} {Deep research system card}.
\newblock Technical report, OpenAI.

\bibitem[{Pan et~al.(2025)Pan, Cemri, Agrawal, Yang, Chopra, Tiwari, Keutzer, Parameswaran, Ramchandran, Klein, Gonzalez, Zaharia, and Stoica}]{pan2025why}
Melissa~Z Pan, Mert Cemri, Lakshya~A Agrawal, Shuyi Yang, Bhavya Chopra, Rishabh Tiwari, Kurt Keutzer, Aditya Parameswaran, Kannan Ramchandran, Dan Klein, Joseph~E. Gonzalez, Matei Zaharia, and Ion Stoica. 2025.
\newblock \href {https://openreview.net/forum?id=wM521FqPvI} {Why do multiagent systems fail?}
\newblock In \emph{ICLR 2025 Workshop on Building Trust in Language Models and Applications}.

\bibitem[{Press et~al.(2022)Press, Zhang, Min, Schmidt, Smith, and Lewis}]{press2022measuring}
Ofir Press, Muru Zhang, Sewon Min, Ludwig Schmidt, Noah~A Smith, and Mike Lewis. 2022.
\newblock Measuring and narrowing the compositionality gap in language models.
\newblock \emph{arXiv preprint arXiv:2210.03350}.

\bibitem[{Qin et~al.(2023)Qin, Liang, Ye, Zhu, Yan, Lu, Lin, Cong, Tang, Qian et~al.}]{qin2023toolllm}
Yujia Qin, Shihao Liang, Yining Ye, Kunlun Zhu, Lan Yan, Yaxi Lu, Yankai Lin, Xin Cong, Xiangru Tang, Bill Qian, et~al. 2023.
\newblock Toolllm: Facilitating large language models to master 16000+ real-world apis.
\newblock \emph{arXiv preprint arXiv:2307.16789}.

\bibitem[{Qwen et~al.(2025)Qwen, :, Yang, Yang, Zhang, Hui, Zheng, Yu, Li, Liu, Huang, Wei, Lin, Yang, Tu, Zhang, Yang, Yang, Zhou, Lin, Dang, Lu, Bao, Yang, Yu, Li, Xue, Zhang, Zhu, Men, Lin, Li, Tang, Xia, Ren, Ren, Fan, Su, Zhang, Wan, Liu, Cui, Zhang, and Qiu}]{qwen2025qwen25technicalreport}
Qwen, :, An~Yang, Baosong Yang, Beichen Zhang, Binyuan Hui, Bo~Zheng, Bowen Yu, Chengyuan Li, Dayiheng Liu, Fei Huang, Haoran Wei, Huan Lin, Jian Yang, Jianhong Tu, Jianwei Zhang, Jianxin Yang, Jiaxi Yang, Jingren Zhou, Junyang Lin, Kai Dang, Keming Lu, Keqin Bao, Kexin Yang, Le~Yu, Mei Li, Mingfeng Xue, Pei Zhang, Qin Zhu, Rui Men, Runji Lin, Tianhao Li, Tianyi Tang, Tingyu Xia, Xingzhang Ren, Xuancheng Ren, Yang Fan, Yang Su, Yichang Zhang, Yu~Wan, Yuqiong Liu, Zeyu Cui, Zhenru Zhang, and Zihan Qiu. 2025.
\newblock \href {http://arxiv.org/abs/2412.15115} {Qwen2.5 technical report}.

\bibitem[{Schick et~al.(2023)Schick, Dwivedi-Yu, Dess{\`\i}, Raileanu, Lomeli, Hambro, Zettlemoyer, Cancedda, and Scialom}]{schick2023toolformer}
Timo Schick, Jane Dwivedi-Yu, Roberto Dess{\`\i}, Roberta Raileanu, Maria Lomeli, Eric Hambro, Luke Zettlemoyer, Nicola Cancedda, and Thomas Scialom. 2023.
\newblock Toolformer: Language models can teach themselves to use tools.
\newblock \emph{Advances in Neural Information Processing Systems}, 36:68539--68551.

\bibitem[{Song et~al.(2025)Song, Jiang, Min, Chen, Chen, Zhao, Fang, and Wen}]{song2025r1}
Huatong Song, Jinhao Jiang, Yingqian Min, Jie Chen, Zhipeng Chen, Wayne~Xin Zhao, Lei Fang, and Ji-Rong Wen. 2025.
\newblock R1-searcher: Incentivizing the search capability in llms via reinforcement learning.
\newblock \emph{arXiv preprint arXiv:2503.05592}.

\bibitem[{Team et~al.(2025)Team, Du, Gao, Xing, Jiang, Chen, Li, Xiao, Du, Liao et~al.}]{team2025kimi}
Kimi Team, Angang Du, Bofei Gao, Bowei Xing, Changjiu Jiang, Cheng Chen, Cheng Li, Chenjun Xiao, Chenzhuang Du, Chonghua Liao, et~al. 2025.
\newblock Kimi k1.5: Scaling reinforcement learning with llms.
\newblock \emph{arXiv preprint arXiv:2501.12599}.

\bibitem[{Trivedi et~al.(2022)Trivedi, Balasubramanian, Khot, and Sabharwal}]{trivedi2021musique}
Harsh Trivedi, Niranjan Balasubramanian, Tushar Khot, and Ashish Sabharwal. 2022.
\newblock {M}u{S}i{Q}ue: Multihop questions via single-hop question composition.
\newblock \emph{Transactions of the Association for Computational Linguistics}.

\bibitem[{Verma et~al.(2025)Verma, Midigeshi, Sinha, Solin, Natarajan, and Sharma}]{verma2025planragefficienttesttimeplanning}
Prakhar Verma, Sukruta~Prakash Midigeshi, Gaurav Sinha, Arno Solin, Nagarajan Natarajan, and Amit Sharma. 2025.
\newblock \href {http://arxiv.org/abs/2410.20753} {Plan*rag: Efficient test-time planning for retrieval augmented generation}.

\bibitem[{Wang et~al.(2024{\natexlab{a}})Wang, Wang, Gao, Zhang, Wu, Xu, Shi, Wang, Li, Qian et~al.}]{wang2024searching}
Xiaohua Wang, Zhenghua Wang, Xuan Gao, Feiran Zhang, Yixin Wu, Zhibo Xu, Tianyuan Shi, Zhengyuan Wang, Shizheng Li, Qi~Qian, et~al. 2024{\natexlab{a}}.
\newblock Searching for best practices in retrieval-augmented generation.
\newblock In \emph{Proceedings of the 2024 Conference on Empirical Methods in Natural Language Processing}, pages 17716--17736.

\bibitem[{Wang et~al.(2024{\natexlab{b}})Wang, Yuan, Dong, Cong, and Li}]{wang2024corag}
Ziting Wang, Haitao Yuan, Wei Dong, Gao Cong, and Feifei Li. 2024{\natexlab{b}}.
\newblock Corag: A cost-constrained retrieval optimization system for retrieval-augmented generation.
\newblock \emph{arXiv preprint arXiv:2411.00744}.

\bibitem[{{xAI}(2025)}]{xai2025grok}
{xAI}. 2025.
\newblock \href {https://x.ai/news/grok-3} {Grok 3}.

\bibitem[{Yang et~al.(2024)Yang, Chern, Qiu, Neubig, and Liu}]{yang2024alignment}
Yuqing Yang, Ethan Chern, Xipeng Qiu, Graham Neubig, and Pengfei Liu. 2024.
\newblock Alignment for honesty.
\newblock \emph{Advances in Neural Information Processing Systems}, 37:63565--63598.

\bibitem[{Yang et~al.(2018)Yang, Qi, Zhang, Bengio, Cohen, Salakhutdinov, and Manning}]{yang2018hotpotqa}
Zhilin Yang, Peng Qi, Saizheng Zhang, Yoshua Bengio, William~W. Cohen, Ruslan Salakhutdinov, and Christopher~D. Manning. 2018.
\newblock {HotpotQA}: A dataset for diverse, explainable multi-hop question answering.
\newblock In \emph{Conference on Empirical Methods in Natural Language Processing ({EMNLP})}.

\bibitem[{Yu et~al.(2024)Yu, Zhang, and Feng}]{yu2024auto}
Tian Yu, Shaolei Zhang, and Yang Feng. 2024.
\newblock Auto-rag: Autonomous retrieval-augmented generation for large language models.
\newblock \emph{arXiv preprint arXiv:2411.19443}.

\bibitem[{Yue et~al.(2024{\natexlab{a}})Yue, Yao, Mi, Yu, Yao, and Yu}]{yue2024dots}
Murong Yue, Wenlin Yao, Haitao Mi, Dian Yu, Ziyu Yao, and Dong Yu. 2024{\natexlab{a}}.
\newblock Dots: Learning to reason dynamically in llms via optimal reasoning trajectories search.
\newblock \emph{arXiv preprint arXiv:2410.03864}.

\bibitem[{Yue et~al.(2024{\natexlab{b}})Yue, Zhuang, Bai, Hui, Jagerman, Zeng, Qin, Wang, Wang, and Bendersky}]{yue2024inference}
Zhenrui Yue, Honglei Zhuang, Aijun Bai, Kai Hui, Rolf Jagerman, Hansi Zeng, Zhen Qin, Dong Wang, Xuanhui Wang, and Michael Bendersky. 2024{\natexlab{b}}.
\newblock Inference scaling for long-context retrieval augmented generation.
\newblock \emph{arXiv preprint arXiv:2410.04343}.

\bibitem[{Zheng et~al.(2023)Zheng, Chiang, Sheng, Zhuang, Wu, Zhuang, Lin, Li, Li, Xing, Zhang, Gonzalez, and Stoica}]{DBLP:conf/nips/ZhengC00WZL0LXZ23}
Lianmin Zheng, Wei{-}Lin Chiang, Ying Sheng, Siyuan Zhuang, Zhanghao Wu, Yonghao Zhuang, Zi~Lin, Zhuohan Li, Dacheng Li, Eric~P. Xing, Hao Zhang, Joseph~E. Gonzalez, and Ion Stoica. 2023.
\newblock \href {http://papers.nips.cc/paper\_files/paper/2023/hash/91f18a1287b398d378ef22505bf41832-Abstract-Datasets\_and\_Benchmarks.html} {Judging llm-as-a-judge with mt-bench and chatbot arena}.
\newblock In \emph{Advances in Neural Information Processing Systems 36: Annual Conference on Neural Information Processing Systems 2023, NeurIPS 2023, New Orleans, LA, USA, December 10 - 16, 2023}.

\bibitem[{Zheng et~al.(2024)Zheng, Sun, Qiu, Ru, Jiayang, Li, Lin, Wang, Luo, Pan, Xu, Min, Zhang, Wang, Li, and Liu}]{zheng-etal-2024-openresearcher}
Yuxiang Zheng, Shichao Sun, Lin Qiu, Dongyu Ru, Cheng Jiayang, Xuefeng Li, Jifan Lin, Binjie Wang, Yun Luo, Renjie Pan, Yang Xu, Qingkai Min, Zizhao Zhang, Yiwen Wang, Wenjie Li, and Pengfei Liu. 2024.
\newblock \href {https://doi.org/10.18653/v1/2024.emnlp-demo.22} {{O}pen{R}esearcher: Unleashing {AI} for accelerated scientific research}.
\newblock In \emph{Proceedings of the 2024 Conference on Empirical Methods in Natural Language Processing: System Demonstrations}, pages 209--218, Miami, Florida, USA. Association for Computational Linguistics.

\end{thebibliography}

\appendix

\section{Prompts}

\subsection{Prompt for Question Quality Level Evaluation}
\label{appendix:prompt_quality_control}

The prompt below displays two templates. Identifies if questions are time-sensitive, subjective, or potentially harmful. Includes classification guidelines, question placeholder, and required answer tag format.

\begin{tcolorbox}[colframe=black, title={Prompt for training data quality checking}]
Please identify whether the given question is time-sensitive, subjective, or may cause harmful answers.\\\\
- Time-sensitive: The answer to the question may change over time.\\
- Harmful: The answer to the question may be harmful or offensive.\\
- Subjective: The answer to the question may be subjective and not based on facts.\\\\
Here is the question:\\
$<$question$>$\\
\{question\}\\
$<$/question$>$\\\\
Wrap your answer in $<$answer$>$ tags with one of the following values:\\
- time\_sensitive: if the question is time-sensitive\\
- harmful: if the question may cause harmful answers\\
- subjective: if the question is subjective\\
- good: if the question is none of the above

\end{tcolorbox}

The prompt below shows the template prompt for contamination detection. To tests if AI responses are influenced by training data contamination.

\noindent\begin{tcolorbox}[colframe=black, title={Prompt for contamination detection}]

Give a short answer to the following question. The answer should be in English.
\\

Question: \{question\}
\\

Your answer:
\end{tcolorbox}

\subsection{Prompt for Model's Answer Quality Level Evaluation}

The prompt below provides instructions for evaluating the correctness of AI-generated answers (pred answer) against a list of ground truth answers. To judge if a predicted answer correctly answers a question by comparing it to ground truth answers.

\label{emb prompt}
\begin{tcolorbox}[colframe=black, title={Prompt for Model-based Evaluation}]
You will be given a question and its ground truth answer list where each item can be a ground truth answer. Provided a pred\_answer, you need to judge if the pred\_answer correctly answers the question based on the ground truth answer list. \\
You should first give your rationale for the judgement, and then give your judgement result (i.e., correct or incorrect). \\\\
Here is the criteria for the judgement: \\
1. The pred\_answer doesn't need to be exactly the same as any of the ground truth answers, but should be semantically same for the question. \\
2. Each item in the ground truth answer list can be viewed as a ground truth answer for the question, and the pred\_answer should be semantically same to at least one of them. \\\\
question: \{question\} \\
ground truth answers: \{gt\_answer\} \\
pred\_answer: \{pred\_answer\} \\\\
The output should in the following json format: \\
'''json \\
\{ \\
"rationale": "your rationale for the judgement, as a text", \\
"judgement": "your judgement result, can only be 'correct' or 'incorrect'" \\
\} \\
''' \\\\
Your output:

\end{tcolorbox}

\subsection{Prompt for Research Plan on Question Answering}

The prompt below outlines the structured approach for addressing complex questions, utilizing web search and webpage browsing tools to conduct in-depth research and gather the necessary information for a comprehensive response.

\label{emb prompt}
\begin{tcolorbox}[colframe=black, title={Prompt for Research Plan on Complex Question Answering}]
\#\# Background information  \\
* Today is YYYY-MM-DD \\
* You are Deep AI Research Assistant \\
The question I give you is a complex question that requires a *deep research* to answer.\\

I will provide you with two tools to help you answer the question:\\
* A web search tool to help you perform google search. \\
* A webpage browsing tool to help you get new page content.\\
\\
You don't have to answer the question now, but you should first think about the research plan or what to search next.\\

Your output format should be one of the following two formats:\\

\texttt{<think>}\\
YOUR THINKING PROCESS\\
\texttt{</think>}
\\
\texttt{<answer>}\\
YOUR ANSWER AFTER GETTING ENOUGH INFORMATION\\
\texttt{</answer>}

or

\texttt{<think>}\\
YOUR THINKING PROCESS\\
\texttt{</think>}\\
\texttt{<tool\_call>}\\
YOUR TOOL CALL WITH CORRECT FORMAT\\
\texttt{</tool\_call>}\\

You should always follow the above two formats strictly.\\
Only output the final answer (in words, numbers or phrase) inside the \texttt{<answer>}\texttt{</answer>} tag, without any explanations or extra information. If this is a yes-or-no question, you should only answer yes or no.\\

\# Tools

You may call one or more functions to assist with the user query.

You are provided with function signatures within \texttt{<tools>}\texttt{</tools>} XML tags:\\
\texttt{<tools>}\\
\{'type': 'function', 'function': \{'name': 'web\_search', 'description': 'Search the web for relevant information from google. You should use this tool if the historical page content is not enough to   answer the question. Or last search result is not relevant to the question.', 'parameters': \{'type': 'object', 'properties': \{'query': \{'type': 'array', 'items': \{'type': 'string', 'description': 'The query to search, which helps answer the question'\}, 'description': 'The queries to search'\}\}, 'required': ['query'], 'minItems': 1, 'uniqueItems': true\}\}\}\\
\{'type': 'function', 'function': \{'name': 'browse\_webpage', 'description': 'Browse the webpage and return the content that not appeared in the conversation history. You should use this tool if the last action is search and the search result maybe relevant to the question.', 'parameters': \{'type': 'object', 'properties': \{'url\_list': \{'type': 'array', 'items': \{'type': 'string', 'description': 'The chosen url from the search result, do not use url that not appeared in the search result'\}, 'description': 'The chosen urls from the search result.'\}\}, 'required': ['url\_list']\}\}\}\\
\texttt{</tools>}\\

For each function call, return a json object with function name and arguments within \texttt{<tool\_call>}\texttt{</tool\_call>} XML tags:\\
\texttt{<tool\_call>}\\
{"name": \texttt{<function-name>}, "arguments": \texttt{<args-json-object>}}\\
\texttt{</tool\_call>}

\end{tcolorbox}

\section{Training Scaling Result}

\begin{figure}[h]
    \centering
    
    \includegraphics[width=\textwidth,page=1]{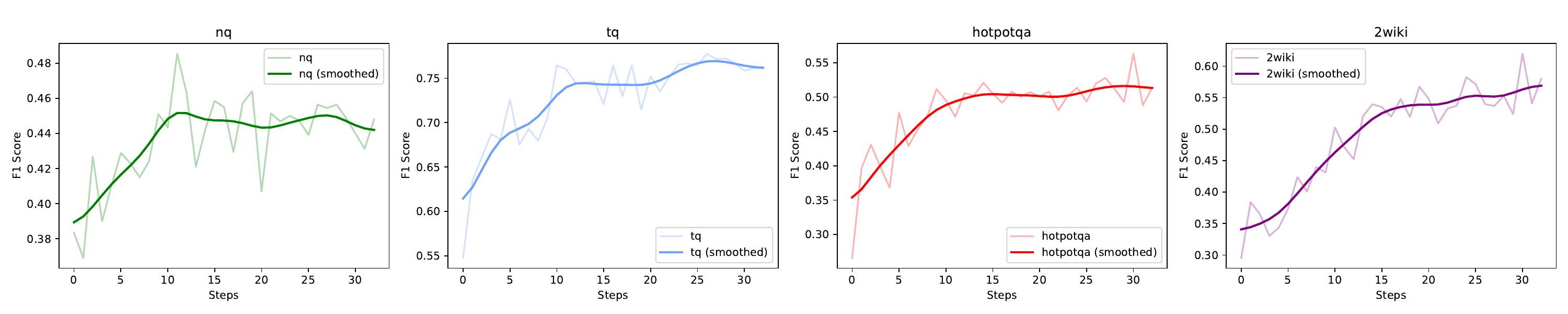}
    
    \includegraphics[width=\textwidth,page=1]{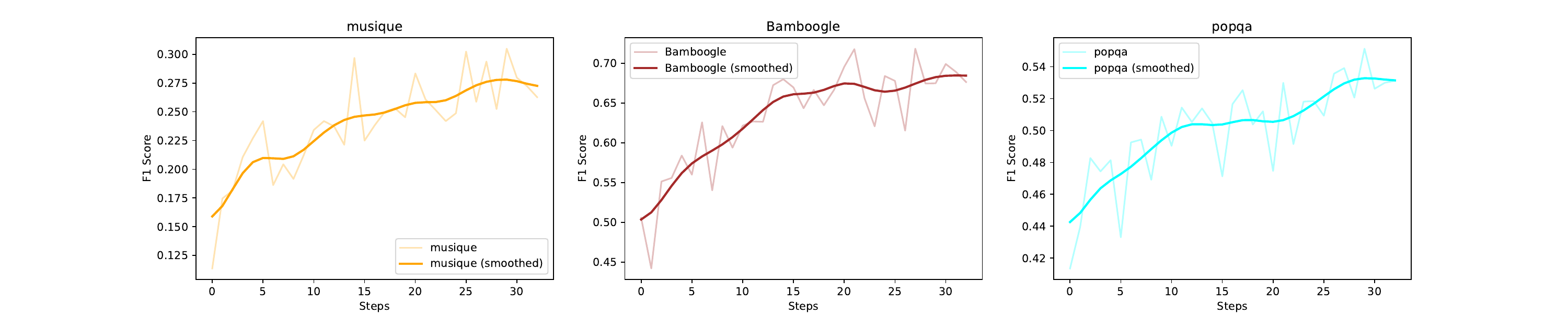}
    
    \caption{F1 score during training}
    \label{fig:f1}
\end{figure}

Figure \ref{fig:f1} presents the F1 score in 7 benchmarks. We sampled 125 cases from each benchmarks' development set. \scrl can scale in all benchmarks, especially in OOD benchmarks.

\end{document}